%% file: main.tex
\documentclass{article}

\usepackage{microtype}
\usepackage{graphicx}
\usepackage{subcaption}
\usepackage{booktabs} 
\usepackage{enumitem}
\usepackage{xcolor}
\definecolor{cvprblue}{rgb}{0.21,0.49,0.74}
\usepackage{hyperref}

\usepackage[preprint]{icml2026}

\usepackage{amsmath}
\usepackage{amssymb}
\usepackage{mathtools}
\usepackage{amsthm}

\input{math_commands}

\usepackage[capitalize,noabbrev]{cleveref}

\theoremstyle{plain}

\theoremstyle{definition}

\theoremstyle{remark}

\usepackage[textsize=tiny]{todonotes}

\usepackage{multirow}     %
\usepackage[table]{xcolor}  %

\definecolor{line-blue1}{HTML}{E7F2FA}
\definecolor{line-blue2}{HTML}{D7E1E8}
\definecolor{line-blue3}{HTML}{D0E5F5}
\definecolor{linecolor_green}{HTML}{B2D2B2} %
\definecolor{linecolor2}{RGB}{246, 248, 239}
\definecolor{linecolor1}{RGB}{230, 234, 217}
\definecolor{linecolor5}{HTML}{4C7C31} %
\definecolor{lightgray}{gray}{0.95}   %

\definecolor{linecolor3}{HTML}{CECEE1} %
\definecolor{linecolo42}{HTML}{B2D2B2} %

\definecolor{linecolor_box}{HTML}{F5F1F8}
\definecolor{linecolor_bla}{HTML}{EBE4F2}
\newcommand{\rotlabel}[1]{\rotatebox{0}{\parbox{2.6cm}{\centering\textbf{#1}}}}
\newcommand{\benchmark}{\texttt{VIA-Bench}\xspace}
\newcommand{\custompara}[1]{{\vspace{1mm}\noindent\textbf{#1}\xspace}}

\usepackage{tcolorbox}

\definecolor{color_blue}{HTML}{F7FDFE} %
\definecolor{color_dark_blue}{HTML}{C7E8F0} %

\newtcolorbox{formattedquote}{
    colback=color_blue,
    colframe=color_dark_blue,
    fontupper=\footnotesize,
    boxsep=-2pt %
}

\usepackage{fvextra} 
\RecustomVerbatimEnvironment{Verbatim}{Verbatim}{
  breaklines=true,      
  breakanywhere=true,  
  breaksymbolleft=\tiny\(\hookrightarrow\)\, 
}

\usepackage{tcolorbox}
\tcbuselibrary{skins} %

\newcommand{\infobox}[1]{
    \vspace{-0.18cm}
    \begin{tcolorbox}[
        colback=linecolor2,     
        colframe=linecolor_green,   
        arc=5pt,                   
        boxsep=5pt,                 
        left=5pt,                  
        right=10pt,                 
        top=2pt,                   
        bottom=3pt,                
        boxrule=0.8pt,              
        drop shadow=linecolor1, 
        enhanced jigsaw             
    ]
    \vspace{-0.1cm}
         \textit{#1}
    \vspace{-0.2cm}
    \end{tcolorbox}
    \vspace{-0.15cm}
}

\newcommand{\Geminilogo}{\raisebox{-1pt}{\includegraphics[scale=0.5]{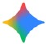}}\xspace}
\newcommand{\Qwenlogo}{\raisebox{-1pt}{\includegraphics[scale=0.5]{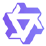}}\xspace}
\newcommand{\Ofourlogo}{\raisebox{-1pt}{\includegraphics[scale=0.08]{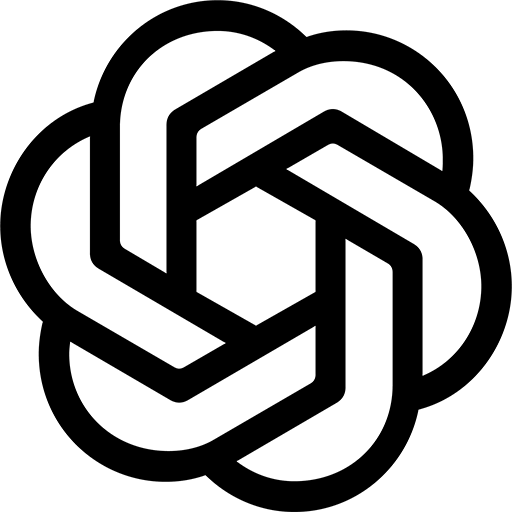}}}

\usepackage{caption} 
\usepackage{xspace}

\makeatletter
\DeclareRobustCommand\onedot{\futurelet\@let@token\@onedot}
\def\@onedot{\ifx\@let@token.\else.\null\fi\xspace}

\def\eg{\emph{e.g}\onedot} 
\def\ie{\emph{i.e}\onedot}

\makeatother
\icmltitlerunning{Seeing Is Believing? A Benchmark for Multimodal Large Language Models on Visual Illusions and Anomalies}

\begin{document}

\twocolumn[
  \icmltitle{Seeing Is Believing? A Benchmark for Multimodal Large Language Models on \texorpdfstring{\\}{ } Visual Illusions and Anomalies}

  \icmlsetsymbol{equal}{*}
  \icmlsetsymbol{intern}{$\dagger$}

  \begin{icmlauthorlist}
       \icmlauthor{Wenjin Hou}{zju,intern}
    \icmlauthor{Wei Liu}{tencent}
    \icmlauthor{Han Hu}{tencent}
    \icmlauthor{Xiaoxiao Sun}{stanford}
    \icmlauthor{Serena Yeung-Levy}{stanford}
    \icmlauthor{Hehe Fan}{zju}
  \end{icmlauthorlist}

  \icmlaffiliation{zju}{Zhejiang University}
  \icmlaffiliation{tencent}{Tencent Hunyuan Team}
  \icmlaffiliation{stanford}{Stanford University}
  
  \icmlcorrespondingauthor{Hehe Fan}{hehefan@zju.edu.cn}

  \icmlkeywords{Machine Learning, ICML}

  \vskip 0.3in

  \begin{center}
    \centering
    \begin{minipage}{\textwidth}
        \centering

        {\includegraphics[width=0.85\linewidth]{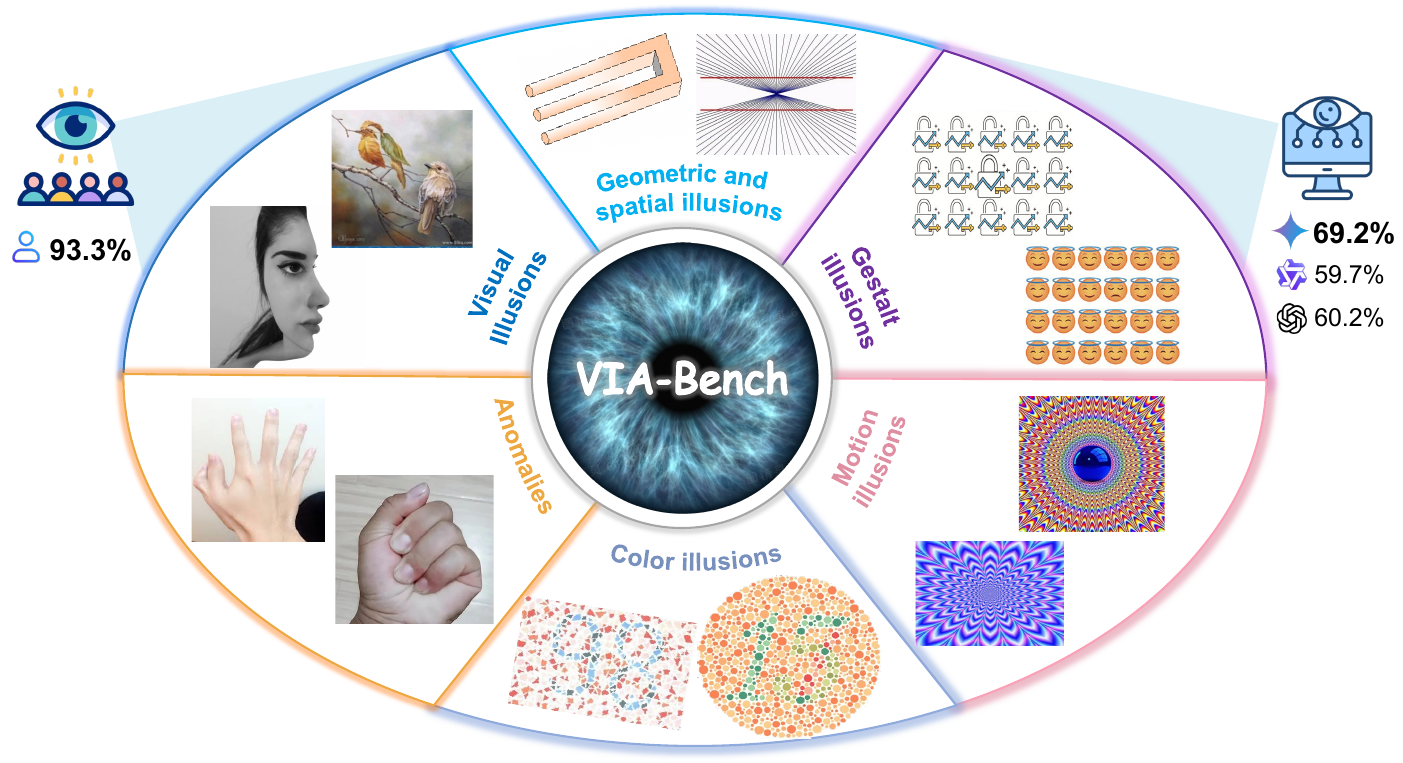}}
        \vspace{-2mm}
        \captionof{figure}{Overview of \benchmark. The benchmark includes six categories: color illusions, motion illusions, gestalt illusions, geometric and spatial illusions, general visual illusions, and visual anomalies. These scenarios require MLLMs to develop human-like perception and deliberate reasoning. On \benchmark, humans achieve 93.30\% average accuracy, whereas the best MLLM reaches 69.23\%.}
        \label{fig:teaser}
    \end{minipage}
  \end{center}

  \vskip 0.1in
]

\printAffiliationsAndNotice{\textsuperscript{$\dagger$}Work done during an internship at Tencent Hunyuan Team.} 

\thispagestyle{empty}

\input{sec/0_abstract}

\input{sec/1_Introduction}

\input{sec/2_Preliminary}

\input{sec/3_Benchmark}

\input{sec/4_Experiment}

\input{sec/5_Related_Work}
\input{sec/6_Conclusion}

\input{sec/Imapct_statement}
\bibliography{example_paper}
\bibliographystyle{icml2026}

\newpage
\appendix
\onecolumn

\input{sec/X_suppl}

\end{document}

%% file: math_commands.tex
\usepackage{amsmath,amsfonts,bm}

\def\eqref#1{equation~\ref{#1}}

\def\1{\bm{1}}

\DeclareMathAlphabet{\mathsfit}{\encodingdefault}{\sfdefault}{m}{sl}
\SetMathAlphabet{\mathsfit}{bold}{\encodingdefault}{\sfdefault}{bx}{n}

\def\gI{{\mathcal{I}}}

\def\gT{{\mathcal{T}}}



%% file: sec/0_abstract.tex
\begin{abstract}
Multimodal Large Language Models (MLLMs) have shown remarkable proficiency on general-purpose vision-language benchmarks, reaching or even exceeding human-level performance. 
However, these evaluations typically rely on standard in-distribution data, leaving the robustness of MLLMs largely unexamined when faced with scenarios that defy common-sense priors. 
To address this gap, we introduce \benchmark, a challenging benchmark designed to probe model performance on visual illusions and anomalies. 
It includes six core categories: color illusions, motion illusions, gestalt illusions, geometric and spatial illusions, general visual illusions, and visual anomalies.
Through careful human-in-the-loop review, we construct over 1K high-quality question-answer pairs that require nuanced visual reasoning. 
Extensive evaluation of over 20 state-of-the-art MLLMs, including proprietary, open-source, and reasoning-enhanced models, uncovers significant vulnerabilities. 
Notably, we find that Chain-of-Thought (CoT) reasoning offers negligible robustness, often yielding ``brittle mirages'' where the model's logic collapses under illusory stimuli. 
Our findings reveal a fundamental divergence between machine and human perception, suggesting that resolving such perceptual bottlenecks is critical for the advancement of artificial general intelligence. The benchmark data and code will be released.

\end{abstract}

%% file: sec/1_Introduction.tex
\section{Introduction}
\label{sec:intro}

Recent years have witnessed the prominent role of Multimodal Large Language Models (MLLMs) across visual domains \cite{achiam2023gpt, google2025gemini3pro}. 
Frontier proprietary systems have reached human-level performance on a wide spectrum of visual benchmarks~\cite{yue2025mmmu, song2025visualpuzzles}.
Beyond basic perception, these models are evolving into sophisticated reasoning agents \cite{guo2025deepseek, openai:systemcard:o3:o4mini:2025}, exhibiting multi-step Chain-of-Thought (CoT) reasoning and 
``thinking with images'' capabilities \cite{xu2025visual,zhang2025thyme}.

To assess these increasingly powerful capabilities, numerous benchmarks have been proposed from various perspectives.
Most of them, however, still emphasize standard visual contexts and stepwise reasoning \cite{wang2025internvl3,zhao2025chain,zeng2025glm}. 
Parallel efforts probe counterfactual logic or visual puzzles \cite{komanduri2025causalvlbench, song2025visualpuzzles, gao2025pixels}.
Yet, emerging evidence consistently shows that many existing multimodal benchmarks lack sufficient difficulty and discriminative power for frontier models \cite{roberts2025zerobench,wang2025theoretical}, underscoring the gap between superficial benchmark success and genuine visual intelligence.

More precisely, these evaluations rarely stress-test models under \textit{atypical conditions} such as cognitive illusions, perceptual conflicts, and intuition-breaking scenarios, where reliance on canonical priors can directly contradict visual evidence. 
For instance, models often blindly predict ``five'' for a six-fingered hand, prioritizing a learned prototype over what is actually shown (see Fig. \ref{fig:teaser}).
As MLLMs increasingly saturate standard benchmarks \cite{yue2025mmmu}, this gap becomes more consequential and motivates a fundamental question: \textit{Do MLLMs truly perceive visual signals, or are they merely performing sophisticated pattern matching against internalized canonical priors?}

These scenarios (see Fig. \ref{fig:teaser}) are particularly well suited for probing this question as they expose failure modes that standard evaluations often miss and cannot be addressed by superficial statistical shortcuts. 
In other words, such illusions fundamentally challenge the intuition that {\em ``seeing is believing''}.
Dealing with these out-of-distribution situations forces MLLMs to consciously weigh what is seen (visual evidence) against what is known (internal knowledge), fostering human-like perception and deliberate reasoning.

To address these limitations, we present \benchmark, a comprehensive and sufficiently challenging benchmark targeting visual illusions and anomalies. 
As depicted in Fig. \ref{fig:teaser}, \benchmark spans diverse specialized categories: color illusions (CI), motion illusions (MI), gestalt illusions (GI), geometric and spatial illusions (GSI), general visual illusions (VI), and visual anomalies (VA)\textcolor{red}{\footnotemark[1]}\footnotetext[1]{In Appendix \textcolor{cvprblue}{\ref{app:taxonomy}}, we provide detailed descriptions and analyses of these categories.}. 
Although these illusions can mislead humans at first glance, they are ultimately interpretable by human observers. By contrast, state-of-the-art (SOTA) MLLMs still lag substantially behind.

We construct \benchmark via careful human-in-the-loop review. 
For each scene, we focus on two key aspects.
First, image content (the {\em ``What''}): what is the intrinsic nature of the image itself?
Second, the reasoning dimension (the {\em ``How''}): what question should be designed to provoke the model's reasoning capabilities?
To ensure data quality and diversity, we adopt a multi-stage pipeline to generate question–answer (QA) pairs (see Fig. \ref{fig:benchmark_construction}).
After rigorous quality verification, we obtain 1{,}004 multiple-choice questions.

\input{fig/benchmark_construction}

To analyze model behavior, we conduct extensive evaluations, including 6 open-source, 7 proprietary systems, and 11 reasoning-enhanced models, with parameter counts ranging from 3B to 235B.
We perform a preliminary analysis and distill three core insights. 
{\em First}, advanced MLLMs still exhibit a major bottleneck compared to humans (see Table \ref{tab:main_results}). Our \benchmark serves as a rigorous testbed to quantify this gap.
{\em Second}, widely used CoT reasoning often degrades accuracy on this benchmark (see Table \ref{tab:cot_results}). This quantifies the models' robustness and vulnerability.
{\em Third}, inconsistent performance across categories further exposes underlying limitations (see Table \ref{tab:main_results} and Fig. \ref{fig:example_evaluation}). With this work, we aim to incentivize the community to explore the headroom of MLLMs and illuminate directions.
This work highlights the following contributions:

{\begin{itemize}[leftmargin=*]
    \item \textbf{The \benchmark Dataset:} We develop a challenging benchmark targeting visual illusions and anomalies, comprising six specialized categories and 1{,}004 meticulously curated QA pairs. 
    \item \textbf{Extensive Benchmarking:} We evaluate 20+ MLLMs, including proprietary systems, open-source models, and reasoning-enhanced architectures. Our results reveal that even the most advanced models lag substantially behind human performance, exposing a persistent bottleneck in machine perception. 
    \item \textbf{The CoT Paradox:} Through detailed analysis, we reveal that CoT reasoning often provides little robustness against these scenarios. This finding suggests that current reasoning mechanisms may amplify internalized priors, offering critical insights for MLLM development.
\end{itemize}

%% file: fig/benchmark_construction.tex
\begin{figure*}[htbp]
    \centering
       \includegraphics[width=1.0\linewidth]
    {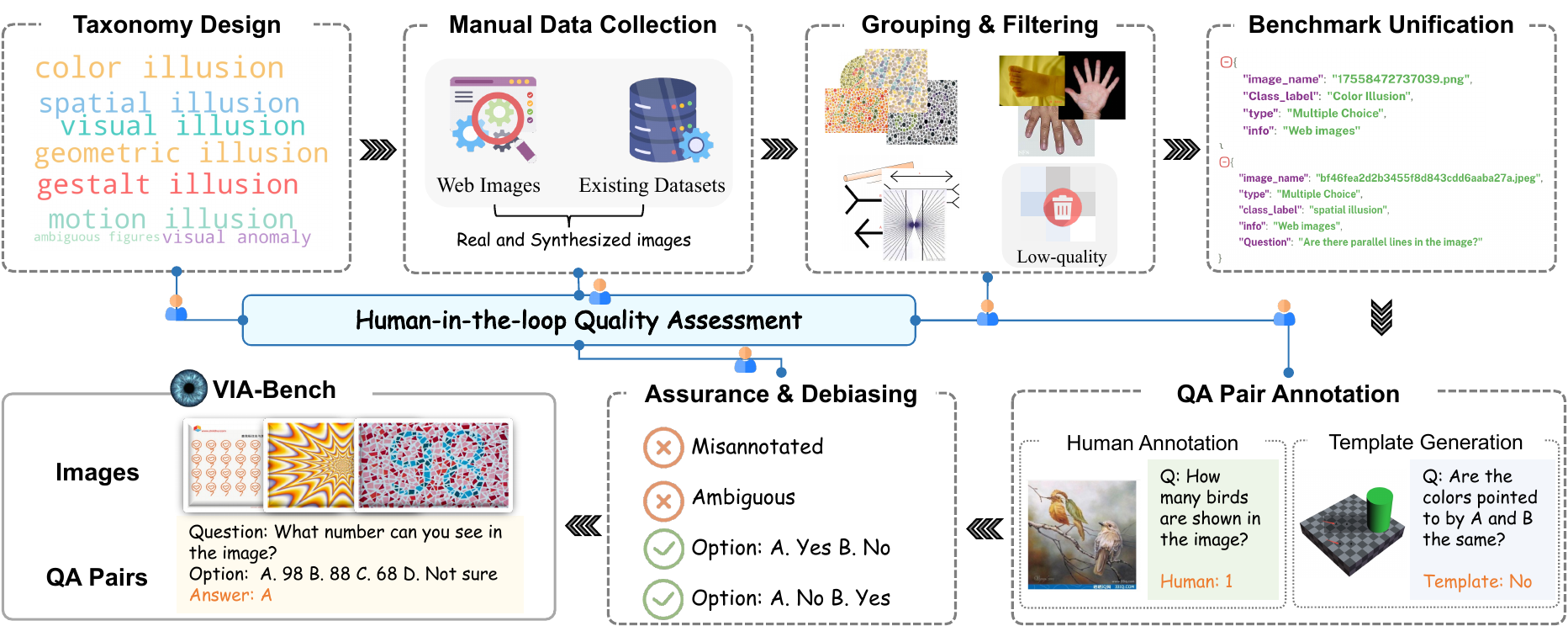}
    \caption{Benchmark construction pipeline. The workflow progresses from data collection to unification, annotation, and debiasing, ultimately forming \benchmark. To ensure high quality, we apply human-in-the-loop assessment at all key stages.}

    \label{fig:benchmark_construction}
\end{figure*}

%% file: sec/2_Preliminary.tex
\section{Preliminaries and Problem Formulation}
\label{sec:preliminaries}
The primary objective of \benchmark is to evaluate the capacity of MLLMs to reconcile conflicting visual stimuli with internal common-sense priors. 
We focus on scenarios where standard reasoning often fails due to the deceptive nature of the visual input. 

\subsection{Formal Task Definition}
Let $\gI$ denote the space of images and $\gT$ denote the space of natural language. For a given instance in \benchmark, we define the input as a tuple $(\mathbf{x}, q, O, i)$, where:
\begin{itemize}
    \item $\mathbf{x} \in \gI$ represents an illusory or anomalous image;
    \item $q \in \gT$ is a task-specific question designed to probe the model's perception;
    \item $O = \{o_1, \dots, o_m\}$ is a set of $m$ candidate options ($m \geq 2$), containing exactly one ground-truth label $y$;
    \item $i \in \gT$ is a formatting instruction (\eg, ``Please provide your answer in the following format: Answer: []'') to ensure deterministic parsing.
\end{itemize}
An MLLM, denoted as a parameterized function $f_\theta$, maps the concatenated visual and textual inputs to a predicted answer $a$: 
\begin{equation*}
    a = f_\theta(\mathbf{x}, q, O, i). 
\end{equation*}
In our evaluation, we further explore the model's reasoning stability by augmenting the instruction $i$ with system-level CoT prompting, which explicitly encourages the model to generate intermediate reasoning steps before selecting $a$.

\subsection{Evaluation Criteria}
To ensure that \benchmark assesses visual intelligence rather than linguistic bias, the question-answer pairs are curated such that the correct option $y$ is statistically independent of the textual priors in $q$. Specifically, the ground truth is established based on the intrinsic properties of $\mathbf{x}$. Success in \benchmark requires the model to transcend canonical priors (\eg, ``a hand typically has five fingers'') in favor of precise visual evidence (\eg, ``this specific hand has six fingers'').

%% file: sec/3_Benchmark.tex
\section{The \benchmark Dataset}

We introduce \benchmark, a diagnostic evaluation suite designed to probe the perceptual robustness of MLLMs. The dataset consists of 1,004 high-quality question-answer (QA) pairs, meticulously curated to isolate failure modes where internal model priors conflict with raw visual evidence.

\subsection{Dataset Construction Pipeline}

As illustrated in Fig. \ref{fig:benchmark_construction}, we employ a systematic, multi-stage, human-in-the-loop pipeline to ensure consistently high-quality and diverse data.

\textbf{Taxonomy and Categorization.} Our taxonomy is grounded in established cognitive psychology literature \cite{eagleman2001visual} and informed by a preliminary failure-mode analysis of frontier MLLMs \cite{comanici2025gemini}. We categorize instances into six primary domains: color (CI), motion (MI), Gestalt (GI), geometric/spatial (GSI), general visual illusions (VI), and visual anomalies (VA). While categories CI through GSI target specific fine-grained perceptual triggers, VA focuses on broader common-sense violations (\eg, structural impossibilities) that challenge the model’s internalized world model.

\input{fig/statistics}

\textbf{Data Curation and Unification.} We utilize a hybrid acquisition strategy, combining manual web-crawling from specialized public sources with the aggregation of high-resolution perceptual datasets (\eg, Turing Eye Test (TET) \cite{gao2025pixels}). Each candidate image underwent a rigorous cleaning process, filtered based on resolution, clarity, and non-ambiguity. To facilitate scalable evaluation, all metadata is unified into a structured schema before the annotation phase.

\textbf{QA Pair Annotation.} We formulate \benchmark as a multiple-choice task to allow for deterministic evaluation. Annotation focuses on three critical dimensions:

\textit{ 1) Inquisitive Precision:} Questions are tailored to specific illusion types instead of being generic (\eg, ``What number is embedded in the pattern?'' for CI). For spatial comparisons, we provide localized visual cues like arrows or letter labels to minimize grounding errors.

\textit{2) Distractor Design:} We manually craft plausible distractors that encapsulate common machine-learning shortcuts or stereotypical priors. Crucially, we include a ``Not Sure'' option in every set to gauge model uncertainty and overconfidence, though it is never the ground-truth label.

\textit{3) Ground-Truth Verification:} All annotations undergo cross-verification by a secondary expert annotator to ensure the ``intrinsic truth'' of the image remains undisputed. 

\textbf{Quality Assurance and Debiasing.} To mitigate shortcut learning, we apply two primary debiasing techniques: (1) randomizing the permutation of options to eliminate positional bias, and (2) for binary questions, randomly flipping question polarity (\eg, alternating between ``Is X true?'' and ``Is X false?'') to ensure that correctness does not correlate with label frequency.

\subsection{Dataset Statistics}
Fig. \ref{fig:statistics} shows the statistical summary of \benchmark. 
    
    \noindent \textbf{Distribution:} As shown in Fig. \ref{fig:statistics}(a), the dataset maintains a balanced distribution across categories, with VA (23.8\%) and GI (11.7\%) representing the upper and lower bounds.
    
    \noindent \textbf{Capability Coverage:} The benchmark spans seven capability dimensions (Fig. \ref{fig:statistics}(b)), testing the intersection of low-level perception and high-level reasoning. 
    
    \noindent \textbf{Structural Diversity:} Question lengths (Fig. \ref{fig:statistics}(c)) vary significantly, reflecting the complexity of the required reasoning. Furthermore, the inclusion of ultra-high-resolution images (up to $8334 \times 2501$) ensures that the benchmark remains challenging for current and future SOTA models.

\input{tab/comparison}

%% file: fig/statistics.tex

\begin{figure}[!t]
    \centering
    \includegraphics[width=1\linewidth]{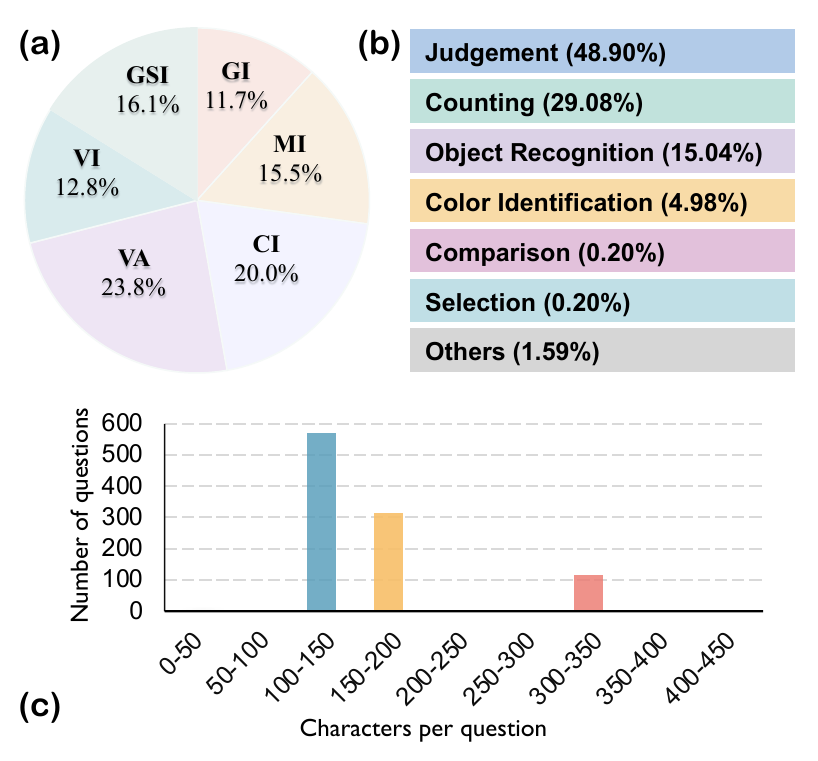}
\caption{Statistical characterization of \benchmark. \textbf{(a)} Distribution across the six primary categories of illusions and anomalies. \textbf{(b)} Mapping of the dataset to specific perceptual and cognitive capabilities.  \textbf{(c)} Empirical distribution of question lengths.}

    \label{fig:statistics}
    \vspace{-4mm}
\end{figure}

%% file: tab/comparison.tex
\begin{table*}[t!]
\centering
\caption{Evaluation on \benchmark. We employ two evaluation protocols--Pattern Match (\ie, Match) and LLM-as-a-Judge (\ie, Judge)--to assess 20+ MLLMs across six categories. They are grouped into proprietary, open-source, and reasoning-enhanced models for the experiment. Each model is run 5 times, and the results are averaged to reduce randomness. Within each group, \raisebox{0pt}[0pt][0pt]{\colorbox{linecolor1}{dark green}} indicates the best result and \raisebox{0pt}[0pt][0pt]{\colorbox{linecolor2}{light green}} denotes the second-best result. The results show that no single model performs well across all aspects. }
\vspace{-1mm}
\label{tab:main_results}
\setlength{\tabcolsep}{1.1pt}
\resizebox{1.0\linewidth}{!}{
\begin{tabular}{r@{\hspace{3mm}}cccccccccccccc}
\toprule
\multirow{3}{*}{\begin{tabular}[l]{@{}l@{}}\textbf{Method}\end{tabular}} &
\multirow{3}{*}{\textbf{Avg.}} & \multirow{3}{*}{\textbf{Rank}} & 
\multicolumn{2}{c}{\rotlabel{VA}}  &
\multicolumn{2}{c}{\rotlabel{CI}} &
\multicolumn{2}{c}{\rotlabel{MI}} &
\multicolumn{2}{c}{\rotlabel{GI}} &
\multicolumn{2}{c}{\rotlabel{GSI}} &
\multicolumn{2}{c}{\rotlabel{VI}} \\ \cmidrule(lr){4-5} \cmidrule(lr){6-7} \cmidrule(lr){8-9} \cmidrule(lr){10-11} \cmidrule(lr){12-13} \cmidrule(lr){14-15}
& & &
\text{Match} & \text{Judge} &
\text{Match} & \text{Judge} &
\text{Match} & \text{Judge} &
\text{Match} & \text{Judge} &
\text{Match} & \text{Judge} &
\text{Match} & \text{Judge} \\

\midrule
\rowcolor{lightgray}\multicolumn{15}{c}{\textbf{Blind Evaluation}} \\
       Random Choice & 29.13  & -- 
       & 28.45 & 28.45 
       & 28.36 & 28.36 
       & 32.69 & 32.69 
       & 23.08 & 23.08 
       & 29.63 & 29.63 
       & 32.56 & 32.56\\
       
       GPT-4-Turbo \cite{achiam2023gpt} & 39.61 & --
       & 2.85 & 2.85 
       & 25.87 & 25.87 
       & 87.95 & 87.95
       & 35.04 & 35.04
       & 61.11 & 61.11
       & 24.81 & 24.81\\

\midrule
\rowcolor{lightgray}\multicolumn{15}{c}{\textbf{Proprietary Systems}} \\
    Gemini-3-pro \cite{google2025gemini3pro} & 69.23 & 1
    & \cellcolor{linecolor1}43.51 & \cellcolor{linecolor1}44.49 
    & \cellcolor{linecolor1}53.33 & \cellcolor{linecolor1}56.82 
    & 69.87 & \cellcolor{linecolor1}99.36
    & \cellcolor{linecolor1}79.32 & \cellcolor{linecolor1}83.76
    & 81.23 & 89.88
    & 61.24 & \cellcolor{linecolor1}67.91\\

    OpenAI o4-mini \cite{openai:systemcard:o3:o4mini:2025} & 55.19 & 2
    & 24.02 & 24.02
    & 27.46 & 28.16
    & \cellcolor{linecolor1}94.87 & \cellcolor{linecolor2}94.87
    & 26.84 & 26.84
    & \cellcolor{linecolor1}97.16 & \cellcolor{linecolor1}97.16
    & 60.31 & 61.24\\

    Gemini-2.5-pro \cite{comanici2025gemini} & 55.01 & 3 
    & \cellcolor{linecolor2}40.42 & \cellcolor{linecolor2}38.74 
    & 28.46 & 28.46 
    & 46.54 & 46.54
    & \cellcolor{linecolor2}65.81 & \cellcolor{linecolor2}65.81
    & 83.95 & 83.95 
    & \cellcolor{linecolor1}65.74 & \cellcolor{linecolor2}65.74 \\

    GPT-5-mini \cite{openai2025gpt5} & 49.54 & 4
    & 26.95 & 26.95
    & \cellcolor{linecolor2}49.25 & \cellcolor{linecolor2}49.25 
    & 25.13 & 25.13
    & 39.15 & 38.63
    & \cellcolor{linecolor2}95.19 & \cellcolor{linecolor2}95.19
    & \cellcolor{linecolor2}61.86 & 61.86\\

    GPT-4o-2024-11-20 \cite{hurst2024gpt} & 43.44 & 5 
    & 35.31 & 35.31 
    & 29.25 & 29.55 
    & 20.13 & 20.13 
    & 22.39 & 22.39 
    & 90.86 & 90.86 
    & 60.93 & 64.19 \\

    ChatGPT-4o-latest \cite{hurst2024gpt} & 36.61 & 6 
    & 2.34 & 3.85 
    & 11.94 & 12.24
    & 35.51 & 35.51 
    & 26.67 & 16.58
    & 92.10 & 91.85 
    & 55.81 & 54.88 \\

    GPT-5-chat-latest \cite{openai2025gpt5} & 35.23 & 7 
    & 0.25 & 2.26 
    & 7.36 & 8.06 
    & 47.18 & 47.05
    & 17.78 & 26.67
    & 85.54 & 88.15
    & 43.88 & 48.53 \\

\midrule
\rowcolor{lightgray}\multicolumn{15}{c}{\textbf{Open-source Models}} \\
      Qwen-vl-max-latest \cite{Qwen-VL} & 46.81 & 4 
      & \cellcolor{linecolor2}25.52 & \cellcolor{linecolor2}25.52 
      & \cellcolor{linecolor1}34.23 & \cellcolor{linecolor1}33.93 
      & 37.95 & 37.82
      & 43.42 & 42.74
      & 78.89 & 78.89 
      & \cellcolor{linecolor2}61.40 & \cellcolor{linecolor2}61.40 \\


      Qwen2.5-VL-3B-Instruct \cite{bai2025qwen2} & 34.84 & 6 
      & 18.83 & 19.00 
      & 24.58 & 24.58
      & 13.21 & 13.21
      & 32.14 & 32.14 
      & 76.17 & 76.17
      & 44.03 & 44.03 \\

      Qwen2.5-VL-7B-Instruct \cite{bai2025qwen2} & 48.00 & 3 
      & 22.93 & 22.93 
      & 22.09 & 22.09
      & \cellcolor{linecolor2}78.97 & \cellcolor{linecolor2}78.97 
      & 34.53 & 34.19 
      & 76.30 & 76.30
      & 53.33 & 53.33 \\

      Qwen2.5-VL-72B-Instruct \cite{bai2025qwen2} & 49.68 & 2
      & 19.00 & 19.00 
      & \cellcolor{linecolor2}29.15 & \cellcolor{linecolor2}29.15
      & 56.03 & 56.03
      & \cellcolor{linecolor2}48.72 & \cellcolor{linecolor2}48.55 
      & \cellcolor{linecolor2}82.47 & \cellcolor{linecolor2}82.47
      & \cellcolor{linecolor1}62.79 & \cellcolor{linecolor1}62.79 \\

      InternVL3.5-8B (w/o thinking) \cite{wang2025internvl3} & 52.73 & 1 
      & \cellcolor{linecolor1}26.36 & \cellcolor{linecolor1}26.36 
      & 16.92 & 16.92
      & \cellcolor{linecolor1}82.69 & \cellcolor{linecolor1}82.69
      & \cellcolor{linecolor1}52.99 & \cellcolor{linecolor1}52.99
      & \cellcolor{linecolor1}83.95 & \cellcolor{linecolor1}83.95 
      & 53.49 & 53.49 \\

      InternVL3.5-38B (w/o thinking) \cite{wang2025internvl3} & 37.36 & 5  
      & 24.27 & 24.27
      & 21.89 & 21.09
      & 4.49 & 4.49
      & 46.15 & 46.15
      & 73.46 & 74.57
      & 54.26 & 53.18 \\

\midrule
\rowcolor{lightgray}\multicolumn{15}{c}{\textbf{Reasoning-enhanced Models}}\\
    OpenAI o3 \cite{openai:systemcard:o3:o4mini:2025} & 60.20 & 1 
    & 22.01 & \cellcolor{linecolor2}22.01 
    & \cellcolor{linecolor1}51.84 & \cellcolor{linecolor1}51.74
    & 90.51 & 90.51
    & 36.75 & 36.75
    & \cellcolor{linecolor1}92.47 & \cellcolor{linecolor1}92.35
    & \cellcolor{linecolor1}67.75 & \cellcolor{linecolor1}67.75 \\
    
    Claude-opus-4.1-20250805 \cite{anthropic-opus-4.1} & 53.52 & 5
    & 21.84 & 20.67 
    & 31.74 & 31.24 
    & 90.38 & 90.26 
    & 53.16 & 53.16
    & 75.68 & 76.30 
    & 48.99 & 48.84 \\
    
    Claude-sonnet-4-20250514 \cite{anthropic-opus-4.1} & 46.61 & 7
    & 21.09 & 21.09 
    & 23.98 & 22.59
    & 41.40 & 41.28
    & \cellcolor{linecolor1}61.20 & \cellcolor{linecolor1}61.20 
    & 78.64 & 78.64
    & 54.26 & 53.95 \\
    
    Claude-3.5-sonnet-20241022 \cite{anthropic2024claude3} & 39.46 & 9
    & \cellcolor{linecolor1}25.02 & \cellcolor{linecolor1}24.44
    & 28.66 & 28.96
    & 20.38 & 20.38 
    & 46.32 & 46.84 
    & 81.45 & 81.85 
    & 34.81 & 34.47 \\
    
    InternVL3.5-8B (w/ thinking) \cite{wang2025internvl3} & 52.85 & 6 
    & \cellcolor{linecolor2}23.68 & 16.15
    & 33.13 & 20.70
    & \cellcolor{linecolor2}95.90 & 92.95
    & 53.68 & 49.06 
    & 73.46 & 67.28
    & 51.63 & 56.59 \\
    
    InternVL3.5-38B (w/ thinking) \cite{wang2025internvl3} & 37.60 & 10
    & 23.35 & 19.41 
    & 24.58 & 22.09 
    & 20.38 & 18.46 
    & 48.89 & 39.49
    & 72.59 & 69.26
    & 42.95 & 49.77 \\
    
    InternVL3.5-30B-A3B (w/ thinking) \cite{wang2025internvl3} & 54.58 & 4 
    & 19.00 & 15.98 
    & 32.44 & 32.44 
    & 82.82 & 85.00
    & 56.58 & 48.21
    & \cellcolor{linecolor2}86.17 & \cellcolor{linecolor2}84.07 
    & 53.95 & 58.29 \\
    
    GLM-4.5V \cite{hong2025glm} & 45.57 & 8 
    & 21.26 & 19.00 
    & \cellcolor{linecolor2}47.26 & \cellcolor{linecolor2}47.56
    & 13.21 & 12.69
    & 47.01 & 49.57
    & 81.11 & 80.25
    & 62.17 & \cellcolor{linecolor2}65.74 \\

    Qwen3-VL-30B-A3B-Thinking \cite{yang2025qwen3} & 57.07 & 3
    & 18.74 & 15.48
    & 40.10 & 39.90
    & \cellcolor{linecolor1}96.28 & \cellcolor{linecolor1}96.28
    & 45.47 & 45.47
    & 83.83 & 83.09
    & 59.38 & 60.78 \\
    
    Qwen3-VL-235B-A22B-Thinking \cite{yang2025qwen3} & 59.72 & 2
    & 23.51 & 20.08
    & 37.61 & 36.42
    & \cellcolor{linecolor2}95.51 & \cellcolor{linecolor2}95.51
    & \cellcolor{linecolor2}58.29 & \cellcolor{linecolor2}55.56
    & 83.58 & 82.96
    & \cellcolor{linecolor2}62.33 & 65.58\\

\midrule
\rowcolor{lightgray}\multicolumn{15}{c}{\textbf{Human Evaluation}}\\
     Human & 93.30 & --
     & 87.45 & 87.45
     & 85.57 & 85.57
     & 100.00 & 100.00
     & 96.58 & 96.58
     & 98.75 & 98.75
     & 91.47 & 91.47\\
      
\bottomrule
\end{tabular}
}
\end{table*}

%% file: sec/4_Experiment.tex
\section{Evaluation on \benchmark}

\subsection{Benchmark Models}
We conduct an extensive evaluation covering 20+ diverse MLLMs on \benchmark, encompassing proprietary, open-source, and reasoning-enhanced models.
\begin{itemize}
    \item \textbf{Proprietary Systems.} We include Gemini-3-pro \cite{google2025gemini3pro}, Gemini-2.5-pro \cite{comanici2025gemini}, GPT-5-chat-latest \cite{openai2025gpt5}, GPT-4o-2024-11-20 \cite{hurst2024gpt}, ChatGPT-4o-latest \cite{hurst2024gpt}, OpenAI o4-mini \cite{openai:systemcard:o3:o4mini:2025}, GPT-5-mini \cite{openai2025gpt5}. These systems, primarily from OpenAI and Google, represent the current frontier of development.
    \item \textbf{Open-source Models.} Our open-source suite mainly comprises the Qwen-VL and InternVL series, including Qwen-VL-max-latest \cite{Qwen-VL}, Qwen2.5-VL-Instruct (3B, 7B, 72B) \cite{bai2025qwen2}, and InternVL3.5 (without thinking) (8B, 38B) \cite{wang2025internvl3}. These models can be deployed locally and constitute strong open-source baselines.
    \item \textbf{Reasoning-enhanced Models.} Recent models generate long CoT when producing answers, making their thinking process visible. They are typically trained with reinforcement learning to enhance their reasoning ability. Our evaluation recent releases, including OpenAI o3 \cite{openai:systemcard:o3:o4mini:2025}, Claude-Sonnet-4-20250514 \cite{anthropic-opus-4.1}, Claude-3.5-Sonnet-20241022 \cite{anthropic2024claude3}, InternVL3.5-38B (with thinking) \cite{wang2025internvl3}, InternVL3.5-30B-A3B (with thinking) \cite{wang2025internvl3}, GLM-4.5V \cite{hong2025glm}, Qwen3-VL-30B-A3B-Thinking \cite{yang2025qwen3}, and Qwen3-VL-235B-A22B-Thinking \cite{yang2025qwen3}. Where explicit rationales are available, we analyze whether the revealed CoT reflects deliberate reasoning on \benchmark. Note that while all models have reasoning capabilities, our grouping is intended for a relatively fair comparison. 
\end{itemize}

\input{tab/cot}

\subsection{Evaluation Metrics and Details}
To directly reflect model performance in the multiple-choice setting, our evaluation uses standard accuracy: $a = n / N$, where $n$ denotes the number of correctly answered questions and $N$ denotes the total number of questions.
Following prior work \cite{fu2025video, zou2025uni, roberts2025grab}, we include explicit ``output format instructions'' in the prompts so that models return a final option label (\eg, \texttt{Answer: [A/B/C/D]}), ensuring that outputs can be reliably parsed.
To provide comprehensive results, we apply two protocols to compare model responses and ground truth:

    \textbf{Match}: a rule-based evaluation via regular expression matching to extract the final option label and verify it against the gold answer.
    
    \textbf{Judge}: LLM-as-a-Judge \cite{zheng2023judging, chen2024mllm}, in which a LLM serves as the arbiter.

We use GPT-4.1-mini \cite{achiam2023gpt} as our judge model to arbitrate the correctness of the model's responses. 
To reduce randomness, we report the average over five independent runs. We provide results of each run in the Appendix \textcolor{cvprblue}{\ref{app:more_experiment}}. 
The matching rules and judging prompts are detailed in the Appendices \textcolor{cvprblue} {\ref{app:match_pattern}} and \textcolor{cvprblue}{\ref{app:judge_prompts}}. 
For fairness and reproducibility, we set the temperature to $0.8$ where supported, otherwise using models’ default hyperparameters (\eg, OpenAI o3) or the model's recommended parameters (\eg, InternVL3.5). Empirically, higher temperature leads to more diverse model outputs (Table \ref{tab:ablation_tem_results}). 
Most models are accessed via API, whereas selected open-source models are evaluated locally. 
Our evaluation covers models from 3B to 235B parameters.
For locally deployed models, all experiments are conducted on 8 H20 GPUs with tensor parallelism, without quantization.
Model identifiers (\ie, markers) and GitHub links are provided in the Appendix \textcolor{cvprblue}{\ref{app:model_version}}. During evaluation, some API services return no completion tokens if the output is too long. We count such cases as incorrect. For each question, we allow up to three retries.

\subsection{Main Results}
Table \ref{tab:main_results} shows overall model performance on \benchmark. Our key observations are as follows:

\custompara{Blind Evaluation.}
The random chance baseline for \benchmark is an average accuracy of 29.13\%, which is the lower bound. 
However, text-only GPT-4-Turbo \cite{achiam2023gpt} (\ie, with vision disabled) demonstrates surprisingly high accuracy, notably on motion illusions (87.95\%) and geometric and spatial illusions (61.11\%). 
This suggests that, even without visual evidence, the model leverages internal linguistic priors to reason about impossible image states (\eg, a static image depicting motion).

\custompara{Human Performance {\em vs.} MLLMs.}
Human achieves 93.30\% average accuracy on \benchmark. 
In comparison, the best MLLM only reaches an average score of 69.23\%. 
Not surprisingly, human performance surpasses all existing MLLMs by a significant margin of at least 24.07\%, indicating that the capabilities of current models remain far below human level on \benchmark.
This substantial gap underscores that \benchmark remains a non-trivial challenge, exposing the ceiling of current multimodal perception.

\custompara{Proprietary MLLMs.}
Despite a significant performance gap compared to humans, proprietary models achieve human-level results on some aspects. 
For example, o4-mini reaches 97.16\% on geometric and spatial illusions. 
Gemini-3-pro leads in three categories.
Nevertheless, no single model exhibits uniformly strong accuracy across all visual illusions and anomalies.

\input{fig/example_evaluation}

\custompara{Open-source MLLMs.}
While top-tier open-source MLLMs like the Qwen-VL and InternVL families deliver competitive results, their overall performance still lags behind proprietary models.
Moreover, several models (\eg, Qwen2.5-VL-3B-Instruct and InternVL3.5-38B) score even lower than the blind evaluation baseline, indicating unreliable visual perception and reasoning capabilities. 
It is also worth noting that performance on \benchmark does not scale consistently with the model's parameters.

\custompara{Reasoning-enhanced MLLMs.}
In the reasoning-enhanced track, OpenAI o3 and Qwen3-VL-235B-A22B-Thinking  obtain the best and the second-best average accuracy, with a score of 60.2\% and 59.72\%. The ``thinking with images'' feature in OpenAI o3 and enhanced multimodal reasoning in Qwen3-VL-235B-A22B-Thinking enable them to exhibit superior comprehension on \benchmark. However, all reasoning-enhanced models exhibit pronounced weaknesses in visual anomalies and color illusions. The highest observed accuracies are only 25.02\% and 51.84\%, respectively. Furthermore, their overall performance on general visual illusions and gestalt illusions remains suboptimal. In contrast, GLM-4.5V and Claude-3.5-sonnet-20241022 fail on motion illusions. These results indicate that even SOTA reasoning models struggle to handle these challenging visual illusions and anomalies.

\infobox{Overall, across categories, leading MLLMs still fall short of human-like perception and deliberate reasoning on \benchmark. The performance remains non-uniform and brittle.}

\input{fig/cot_visualization}
\subsection{Does CoT Reasoning Help on \benchmark?}
CoT prompting typically enhances complex reasoning \cite{wang2023self,muralidharan2024deliberate}. 
However, recent findings reveal that it can degrade performance on counter-intuitive tasks, where reasoning tends to rationalize deceptive priors rather than correct them \cite{liu2025mind,qin2025chain}.
Motivated by this, we investigate whether text-based CoT aids MLLMs on \benchmark, which is rich in visual perceptual traps that may undermine purely textual step-by-step reasoning.
We evaluate two variants injected via the system prompt: zero-shot CoT (\ie, Let's think step by step) and manual CoT (\ie, providing guides to make the model reason logically). 
The detailed prompt designs are provided in Appendix \textcolor{cvprblue}{\ref{app:system prompt}}.

We compare against a direct prompting baseline in Table \ref{tab:cot_results} and visualized in Fig. \ref{fig:cot_visualization}.
Contrary to general visual tasks, CoT strategies generally fail to yield improvements on \benchmark, often degrading performance.
Qualitative analysis identifies the primary cause as hallucination reinforcement (see Appendix \textcolor{cvprblue}{\ref{app:case study}} for detailed cases).
Instead of correcting the initial perceptual error, the extended reasoning process typically devolves into overthinking.
Since visual illusions trigger an immediate misperception, subsequent textual reasoning steps do not correct the error but instead rationalize it, generating plausible-sounding justifications for the incorrect premise.
A notable exception appears in Qwen2.5-VL-7B, where zero-shot CoT improves accuracy on motion illusions (+14.8\%).
However, closer inspection reveals this stems from invoking \textit{textual priors} (\eg, static image assumptions) rather than genuine visual analysis.
This observation leads to a critical insight:
\infobox{Purely text-based CoT is insufficient for resolving visual illusions. Without continuous visual grounding to verify intermediate steps, CoT tends to either reinforce perceptual errors or rely on ungrounded textual shortcuts derived from priors.}

\subsection{Qualitative Analysis}
To disentangle the primary bottlenecks of frontier MLLMs on our \benchmark, we analyze the response behaviors of OpenAI o3, OpenAI o4, Gemini-2.5-pro, and Qwen3-VL-30B-A3B (see Fig. \ref{fig:example_evaluation}). 
We further conduct a closer inspection of the reasoning traces in Appendix \textcolor{cvprblue}{\ref{app:case study}} (covering  InternVL-3.5-8B, Qwen3-VL-30B-A3B-Thinking, and Gemini-2.5-pro). 
Based on these observations, we derive three key findings.

\custompara{Finding-1:} One clear takeaway is that most error types stem from an initial visual misperception rather than logical fallacies. 
Once the model misidentifies a key visual element (\eg, missing a subtle illusory cue), it creates a flawed premise, rendering the final deduction inevitably incorrect.

\custompara{Finding-2:} On \benchmark, models often exhibit high uncertainty due to the conflict between visual evidence and internalized priors.
Even after successfully extracting visual information, they tend to self-negate (\eg, ``I'm a bit uncertain...'', ``maybe...''), prioritizing canonical knowledge over what is actually presented.

\custompara{Finding-3:} A severe overthinking phenomenon is prevalent \cite{chennot}. Visual illusions trigger repetitive or self-contradictory CoT traces (\eg, loops of ``wait'', ``but'', ``no''), where model struggles to resolve the visual ambiguity.

\input{tab/temperature}

\subsection{Ablation Study}
Table \ref{tab:ablation_tem_results} presents the results of our ablation study across various configurations. We set the temperature to $T=0.8$ for Gemini-2.5-pro and $T=0.6$ for InternVL3.5-8B in our experiments. Results are averaged over five independent trials to ensure statistical significance. 

\vspace{-1mm}

%% file: tab/cot.tex
\begin{table*}[t!]
\centering

\caption{Impact of CoT strategies on \benchmark. We evaluate the performance of Gemini-2.5-pro and Qwen2.5-VL-7B across varying reasoning configurations. Results are averaged over five independent trials to ensure statistical significance. We compare zero-shot performance against manual CoT prompting injected via system instructions. The data reveals that CoT often fails to provide robustness, highlighting the brittle nature of current reasoning mechanisms under illusory stimuli.} 

\label{tab:cot_results}
\setlength{\tabcolsep}{1.1pt}
\resizebox{1.0\linewidth}{!}{
\begin{tabular}{r@{\hspace{3mm}}c@{\hspace{3mm}}cccccccccccc}
\toprule
\multirow{3}{*}{\begin{tabular}[l]{@{}l@{}}\textbf{Method}\end{tabular}} &
\multirow{3}{*}{\textbf{Avg.}} &
\multicolumn{2}{c}{\rotlabel{VA}}  &
\multicolumn{2}{c}{\rotlabel{CI}} &
\multicolumn{2}{c}{\rotlabel{MI}} &
\multicolumn{2}{c}{\rotlabel{GI}} &
\multicolumn{2}{c}{\rotlabel{GSI}} &
\multicolumn{2}{c}{\rotlabel{VI}} \\
\cmidrule(lr){3-4} \cmidrule(lr){5-6} \cmidrule(lr){7-8} \cmidrule(lr){9-10} \cmidrule(lr){11-12} \cmidrule(lr){13-14}
& &
\text{Match} & \text{Judge} &
\text{Match} & \text{Judge} &
\text{Match} & \text{Judge} &
\text{Match} & \text{Judge} &
\text{Match} & \text{Judge} &
\text{Match} & \text{Judge} \\

\midrule
\rowcolor{lightgray}\multicolumn{14}{l}{\textbf{Gemini-2.5-pro}} \\
w/o CoT (\ie, normal) & 55.01
& 40.42 & \cellcolor{linecolor2}38.74
& 28.46 & \cellcolor{linecolor2}28.46
& \cellcolor{linecolor2}46.54 & \cellcolor{linecolor1}46.54
& \cellcolor{linecolor2}65.81 & \cellcolor{linecolor2}65.81
& \cellcolor{linecolor1}83.95 & \cellcolor{linecolor1}83.95
& \cellcolor{linecolor2}65.74 & 65.74 \\

w/ zero-shot CoT & 54.86
& \cellcolor{linecolor2}40.67 & 38.24
& \cellcolor{linecolor2}29.75 & 27.66
& 45.64 & 45.51
& 64.79 & 64.79
& 83.46 & 83.21
& \cellcolor{linecolor1}67.29 & \cellcolor{linecolor1}67.29 \\

w/ manual CoT & 54.32
& \cellcolor{linecolor1}41.09 & \cellcolor{linecolor1}38.49
& \cellcolor{linecolor1}32.34 & \cellcolor{linecolor1}28.76
& 44.23 & \cellcolor{linecolor2}44.23
& \cellcolor{linecolor1}64.79 & \cellcolor{linecolor1}64.79
& \cellcolor{linecolor2}82.22 & \cellcolor{linecolor2}82.22
& 64.34 & \cellcolor{linecolor2}64.34 \\

\rowcolor{lightgray}\multicolumn{14}{l}{\textbf{Qwen2.5-VL-7B}} \\
w/o CoT (\ie, normal) & 48.00
& \cellcolor{linecolor2}22.93 & \cellcolor{linecolor2}22.93
& 22.09 & 22.09
& \cellcolor{linecolor2}78.97 & \cellcolor{linecolor2}78.97
& \cellcolor{linecolor2}34.53 & \cellcolor{linecolor2}34.19
& \cellcolor{linecolor2}76.30 & \cellcolor{linecolor2}76.30
& \cellcolor{linecolor2}53.33 & \cellcolor{linecolor2}53.33 \\

w/ zero-shot CoT & 48.98
& 18.49 & 18.49
& \cellcolor{linecolor2}22.39 & \cellcolor{linecolor2}22.39
& \cellcolor{linecolor1}93.72 & \cellcolor{linecolor1}93.72
& 28.29 & 27.35
& 73.09 & 72.96
& \cellcolor{linecolor1}58.45 & \cellcolor{linecolor1}58.45 \\

w/ manual CoT & 47.10
& \cellcolor{linecolor1}24.77 & \cellcolor{linecolor1}24.77
& \cellcolor{linecolor1}21.00 & \cellcolor{linecolor1}21.00
& 72.56 & 72.56
& \cellcolor{linecolor1}33.16 & \cellcolor{linecolor1}33.16
& \cellcolor{linecolor1}74.81 & \cellcolor{linecolor1}74.81 
& 56.28 & 56.28 \\

\bottomrule
\end{tabular}
}
\end{table*}

%% file: fig/example_evaluation.tex
\begin{figure*}[!t]
    \centering
    \includegraphics[width=1.0\linewidth]{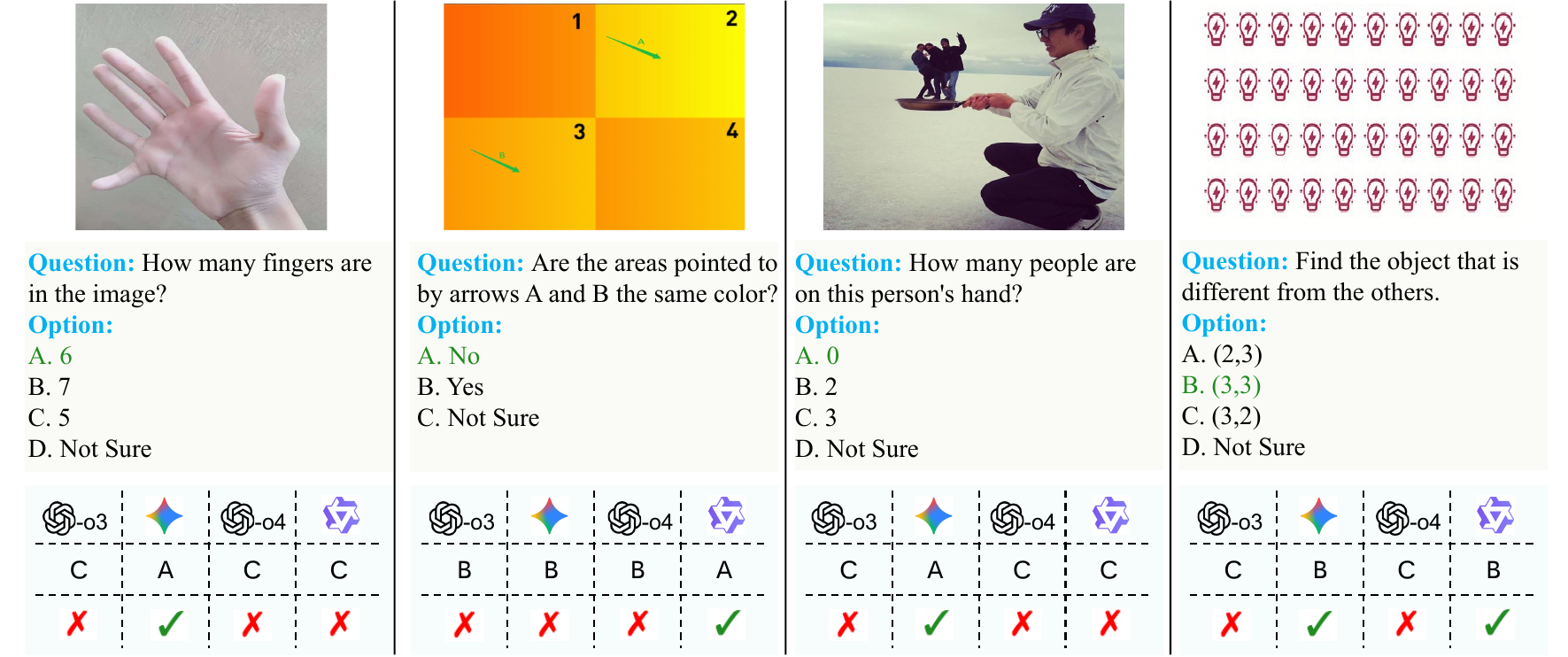}
    \vspace{-6mm}
    \caption{Visualization of model responses
    on \benchmark. These examples demonstrate that even leading models struggle to handle relatively simple tasks such as counting, color recognition, and perceiving fine-grained detail on visual illusions and anomalies.
    \Ofourlogo $\text{-o3}$: OpenAI o3; \Geminilogo: Gemini-2.5-pro; \Ofourlogo $\text{-o4}$: OpenAI o4; \Qwenlogo: Qwen3-VL-30B-A3B-Thinking.
    More cases can be found in the Appendix \textcolor{cvprblue}{H}.}
    \label{fig:example_evaluation}
    \vspace{-2mm}
\end{figure*}

%% file: fig/cot_visualization.tex
\begin{figure}
    \centering
    \includegraphics[width=0.9\linewidth]{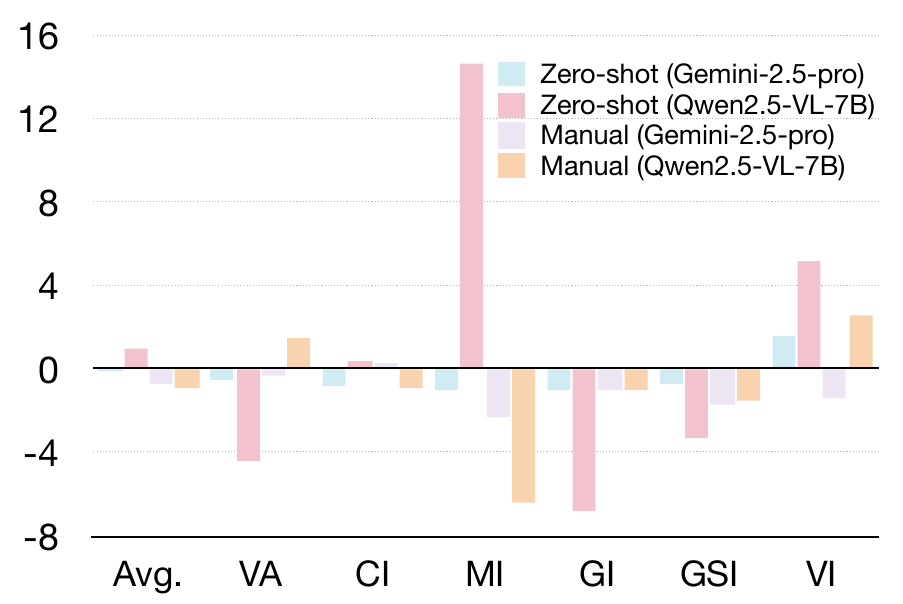}
    \caption{Relative gains of CoT over baseline prompting. Negative values indicate performance attenuation after applying CoT prompts. These results underscore that \benchmark presents a fundamental perceptual bottleneck that is not easily bypassed through surface-level textual prompting.}
    \label{fig:cot_visualization}
    \vspace{-5mm}
\end{figure}

%% file: tab/temperature.tex
\begin{table}[t!]
\centering
\caption{Ablation on the temperature configuration. Rows with \raisebox{0pt}[0pt][0pt]{\colorbox{linecolor1}{dark green}} indicate the values used in our experiments.}
\vspace{-2mm}
\label{tab:ablation_tem_results}
\setlength{\tabcolsep}{1pt}
\resizebox{\linewidth}{!}{ 
\begin{tabular}{l@{\hspace{2mm}}c@{\hspace{2mm}}cccc}
\toprule
\multirow{2}{*}{\begin{tabular}[l]{@{}l@{}}\textbf{Model}\end{tabular}} &
\multirow{2}{*}{\textbf{Tem.}} &
\multicolumn{2}{c}{\rotlabel{VA}}  &
\multicolumn{2}{c}{\rotlabel{CI}} \\
\cmidrule(lr){3-4} \cmidrule(lr){5-6}
& &
\text{Match} & \text{Judge} &
\text{Match} & \text{Judge} \\
\midrule
Gemini-2.5-pro & 0.1 & 40.50 & 40.42 & 30.05 & 30.05 \\
\rowcolor{linecolor2}Gemini-2.5-pro & 0.8 & 40.42 & 38.74 & 29.55 & 26.77 \\
Gemini-2.5-pro & 1.5 & 40.92 & 40.84 & 29.45 & 29.75 \\

InternVL3.5-8B & 0.1 & 23.35 & 14.98 & 34.53 & 20.70\\
\rowcolor{linecolor2}InternVL3.5-8B & 0.6 & 23.68 & 16.15 & 33.13 & 20.70 \\
InternVL3.5-8B & 1.5 & 25.27 & 15.06 & 31.74 & 21.00\\
\bottomrule
\end{tabular}
}
\vspace{-6mm}
\end{table}

%% file: sec/5_Related_Work.tex
\section{Related Works}
\subsection{Multimodal Large Language Models}
Multimodal Large Language Models (MLLMs) have emerged as a foundational paradigm for general-purpose AI.
Advanced proprietary families such as GPT \cite{achiam2023gpt,hurst2024gpt, openai2025gpt5}, the o-series \cite{jaech2024openai,openai:systemcard:o3:o4mini:2025}, Gemini \cite{team2023gemini, comanici2025gemini}, and Claude \cite{anthropic-opus-4.1}, alongside open-source series such as Qwen \cite{Qwen-VL,bai2025qwen2,yang2025qwen3}, InternVL \cite{wang2025internvl3}, LLaVA \cite{liu2023visual}, BLIP \cite{li2023blip,chen2025blip3}, GLM \cite{hong2025glm}, Emu \cite{wang2024emu3}, Show-o \cite{xie2025show} have fueled rapid development.
Recent efforts to enhance their understanding and reasoning capabilities through supervised fine-tuning and reinforcement learning have achieved remarkable performance across various visual benchmarks \cite{guo2025seed1,yu2023mm,chen2024we,liu2024mmbench,sun2024parrot,li2025system,zou2025uni}.
However, studies reveal that MLLMs still struggle to handle cognitive illusions, perceptual conflicts, and intuition-breaking scenarios \cite{gao2025pixels, roberts2025zerobench,rostamkhani2025illusory,li2023trustworthy, chander2025toward}.
This restricts their ability to generalize and be robust in real-world situations~\cite{zitkovich2023rt,kim2024openvla,intelligence2504pi0,cheang2025gr,gong2025space,jia2025omnispatial,liao2025genie,li2025worldeval,team2025gemini,chen2025learning}.

\vspace{-1mm}
\subsection{Multimodal Visual Benchmarks}
With the expanding capabilities of MLLMs, numerous studies have introduced innovative benchmarks to evaluate performance on multimodal tasks.
Broadly, these benchmarks can be categorized into holistic evaluations \cite{yu2023mm, liu2024mmbench, li2023seed,dubey2024llama,wang2025internvl3,yang2025qwen3,zhao2025chain,wang2025theoretical} and domain-specific assessments, such as MathVision \cite{lu2023mathvista}, ChartQA \cite{masry2022chartqa}, OCRBench \cite{liu2023hidden}, and MMMU/MMMU-pro \cite{yue2025mmmu}.
Despite their significance, most prior work focuses on natural-image contexts.
More recently, TET \cite{gao2025pixels} tests four visual perception-to-reasoning tasks. OmniSpatial \cite{jia2025omnispatial} and Space-10 \cite{gong2025space} are grounded in basic spatial relations.
VisualPuzzle \cite{song2025visualpuzzles}, PuzzleVQA \cite{chia2024puzzlevqa}, and AlgopuzzleVQA \cite{ghosal2024language} emphasize the model’s reasoning ability in visual puzzles.
Hallusionbench \cite{guan2024hallusionbench} and HellaSwag-Pro \cite{li2025hellaswag} probe hallucination and counterfactual understanding.
ZeroBench \cite{roberts2025zerobench} and VibeEval \cite{padlewski2024vibe} raise task difficulty to challenge current MLLMs.
A line of related work includes Illusory VQA \cite{rostamkhani2025illusory}, which constructs edited datasets (\eg, IllusionMNIST) for illusion recognition; The Art of Deception \cite{gomez2025art} studies how color visual illusions are encoded in diffusion models; GVIL \cite{zhang2023grounding} probes VLM perception of color and geometric illusions from several root images and tasks. 
Extending beyond prior benchmarks, our \benchmark establishes a systematic, comprehensive testbed for advanced MLLMs, spanning diverse visual illusions and anomalies.

\vspace{-1mm}

%% file: sec/6_Conclusion.tex
\section{Conclusions and Future Outlook}
This paper introduces \benchmark, a comprehensive and challenging benchmark for evaluating frontier MLLMs on visual illusions and anomalies.
\benchmark covers six primary categories.
Through careful human-in-the-loop review, we curated 1{,}004 high-quality multiple-choice QA pairs.
Extensive evaluations of 20+ MLLMs show that even SOTA proprietary systems, open-source, and reasoning-enhanced models peak at 69.23\% accuracy, leaving a 24.07\% gap relative to human performance.
Furthermore, our analysis reveals that CoT reasoning is brittle and often inconsistent on \benchmark.
We also deliver several findings from qualitative case analyses.
By exposing these gaps, we aim to push the boundaries of MLLMs, offering valuable insights for future research toward human-level machine intelligence. 

To further accelerate breakthroughs in the research community, \benchmark will be publicly released.
While the current evaluation covers 20+ MLLMs, the testbed will be continuously updated to include new models as they become available.
Also, more diverse formats, such as open-ended QA, may be iteratively introduced to better challenge perceptual and reasoning abilities in the context of visual illusions.
This may involve adding complex compositional visual illusion questions to \benchmark (\eg, ``do moving objects of different colors but the same length appear in the image?").
Finally, building on the insights from \benchmark, a key research avenue is to improve MLLM generalization in these intuition-challenging edge scenes, potentially by leveraging specific data and appropriate learning objectives. 

\pagebreak

%% file: sec/Imapct_statement.tex
\section*{Impact Statement}
This work introduces \benchmark, a comprehensive benchmark for evaluating MLLMs on visual illusions and anomalies. By exposing critical perceptual failure modes in current systems, \benchmark enhances the understanding of model robustness and reliability. This approach has the potential to advance the development of more trustworthy multimodal AI applications. We do not foresee major ethical or social concerns related to this work.

%% file: sec/X_suppl.tex
\begin{appendix}
\section*{Appendix Outline}
The Appendix is organized as follows.
\begin{itemize}
    \item \textbf{Appendix \ref{app:taxonomy}} details the taxonomy of \benchmark.
    \item \textbf{Appendix \ref{app:question_examples}} provides typical question examples.
    \item \textbf{Appendix \ref{app:more_experiment}} presents additional experiments on  OpenAI o3 and Qwen3-VL-235B-A22B-Thinking.
    \item \textbf{Appendix \ref{app:match_pattern}} describes regular match patterns.
    \item \textbf{Appendix \ref{app:judge_prompts}} lists the input prompt for the judge model.
    \item \textbf{Appendix \ref{app:model_version}} specifies detailed model versions for the 20+ MLLMs.
    \item \textbf{Appendix \ref{app:system prompt}} shows system prompts.
    \item \textbf{Appendix \ref{app:case study}} offers case studies and analysis based on models’ concrete reasoning procedures.
\end{itemize}

\section{Taxonomy of \benchmark}
\label{app:taxonomy}
\benchmark aims to comprehensively evaluate MLLMs under visual illusions and visual anomalies . 
Generally, these scenarios are counterintuitive or countercommonsense for human observers.
Such settings require MLLMs to exhibit human-like perception and deliberate reasoning.
\benchmark covers six major categories: Color Illusions (CI), Motion Illusions (MI), Gestalt Illusions (GI), Geometric and Spatial Illusions (GSI), General Visual Illusions (VI), and Visual Anomalies (VA). 
Each category targets purpose-built reasoning challenges arising from visual illusions or anomalies.
The following presents brief descriptions and representative examples for each category.

\custompara{Color Illusions (CI)}: This category focuses on color constancy, contrast, illumination, and shadow, which lead to color misperception and interference. Typical examples include Checker Shadow illusion, colorblind (Ishihara) plates, and the Bezold effect.

\custompara{Motion Illusions (MI)}: Although the images in this category are static, specific textures and contrasts elicit subjective perceptions of motion or pulsation, such as stripe-induced drift and the spiral illusion.

\custompara{Gestalt Illusions (GI)}: This category involves detecting fine-grained target objects in an image. 
Due to Gestalt effects, perceptual biases can arise from the overall composition. 
A common example is identifying the odd one out within a repetitive pattern.

\custompara{Geometric and Spatial Illusions (GSI)}: This category probes geometric perceptual biases, such as those arising in judgments of length, parallelism, alignment, and 3D existence. 
Examples include impossible figures (\eg, the Penrose triangle), the Zöllner illusion, geometric traps, and the Café Wall illusion.

\custompara{General Visual Illusions (VI)}: Some visual illusions cannot be directly assigned to the categories above. We group these separately as general visual illusions.
This category primarily involves misperceptions and visual misdirection caused by perspective, composition, or occlusion. 
Typical examples include leaves that resemble birds, painting illusions, mirror illusions, and illusion photography.

\custompara{Visual Anomalies (VA)}: This category refers to anomalies that violate commonsense priors. Models are easily misled by prior knowledge and experience when identifying them. 
In this study, we primarily focus on biological anomalies, such as polydactyly (\eg, extra or missing fingers or toes).

\section{Question Examples}
\label{app:question_examples}
To raise the difficulty to levels appropriate to challenge the current and future generations of frontier models, we avoid trivial binary questions such as “Is this a visual illusion?” or “Is this a visual anomaly?” 
Instead, for each category, we design diverse questions from the perspective of a normal image to probe the model's performance on illusions and anomalies. Table~\ref{tab:question_examples} provides several examples.
Furthermore, we construct specific multiple-choice options for each question, including misleading distractors and the ground-truth answer.
This design probes both low-level perception (\eg, counting, color discrimination, fine-grained detail recognition) and higher-level reasoning (\eg, evidence aggregation, consistency checking, and summary judgment) in the presence of illusions and anomalies, offering a comprehensive assessment of model robustness.
\input{tab/templete}

\section{More Experiment Results}
\label{app:more_experiment}
Here, we additionally report the results for each run of OpenAI o3 \cite{openai:systemcard:o3:o4mini:2025} and Qwen3-VL-235B-A22B-Thinking \cite{yang2025qwen3} in Table \ref{tab:num_runs}.  
Although there are variations across runs, our evaluation process--averaging over five runs--substantially reduces model randomness. 
Furthermore, the models' input is formatted as follows: \texttt{[Image][Prompt]}, where \texttt{prompt} includes the question, any available options, and formatting instructions.

\input{tab/num-runs}

\section{Match Patterns}
\label{app:match_pattern}
We use the following regular expression (pattern) to match the answer:
\begin{formattedquote}
\begin{verbatim}
patterns = [
    r'answer\s*[:\-]\s*\[([ABCD])\]',    # Answer: [A] (anywhere in text)
    r'\[([ABCD])\]',                     # [A] (anywhere in text)
    r'answer\s*[:\-]\s*([ABCD])\b',      # Answer: A (with word boundary)  
    r'\b([ABCD])\b',                     # A (single letter with word boundaries)       
    r'answer\s*[:\-]\s*\[([ABCD])\]\b',  # Answer: [A] (with word boundary)
    r'answer\s*is\s*\[([ABCD])\]',       # Answer is [A]
    r'answer\s*is\s*([ABCD])',           # Answer is A
    r'the\s*answer\s*is\s*\[([ABCD])\]', # The answer is [A]
    r'the\s*answer\s*is\s*([ABCD])',     # The answer is A
    r'choice\s*[:\-]\s*\[([ABCD])\]',    # Choice: [A]
    r'option\s*[:\-]\s*\[([ABCD])\]'     # Option: [A]     
]
\end{verbatim}
\end{formattedquote}

\section{Judge Prompts}
\label{app:judge_prompts}
The input prompt for the judge model is in the following format:
\begin{formattedquote}
\begin{Verbatim}
judge_prompt = f"""You are a judge for multiple choice questions.
Analyze this model response and determine what answer it indicates:
Model Response: "{response}"
Based on the content of this response, what option is the model actually choosing?
Return ONLY one letter: A, B, C, or D

Examples:
- "I think it's A" → A
- "The answer should be B" → B
- "My choice is C" → C
- "I'm not sure" → D
- "Cannot determine" → D

What option does this response indicate?"""
\end{Verbatim}   
\end{formattedquote}

\section{Model Versions}
\label{app:model_version}

In Table \ref{tab:model_version}, we provide the detailed model versions for the 20+ MLLMs used in our evaluation. 
For models accessed via API, we report the provider’s unique model identifiers (markers).
For models we ran locally, we include links to the corresponding GitHub repositories.
All models included in our evaluation were available as of November 18, 2025.
\input{tab/model_version}

\section{System Prompts}
\label{app:system prompt}
\input{fig/system_prompt}
Our research indicates that prevailing CoT methods do not effectively handle our \benchmark. 
In this section, we present all system prompts used in our experiments in Fig. \ref{fig:system_prompt}. 
This is intended to showcase the carefully crafted prompts and facilitate reproducibility. 
Unless otherwise specified, the normal system prompt was used in all our experiments.

\section{Case Study}
\label{app:case study}
In this section, we present more case studies for human-conducted analysis of the models' concrete reasoning procedures.
We include three selected frontier models: InternVL-3.5-8B \cite{wang2025internvl3}, Qwen3-VL-30B-A3B-Thinking \cite{yang2025qwen3} and Gemini-2.5-pro \cite{liao2025genie}.
In the analysis, we identify the categorized error types and highlight our findings in relevant parts. Results are shown in Figs. \ref{fig:color_illusion_example}--\ref{fig:visual_anomaly_example}.

For the MLLMs' reasoning processes, we mark correct reasoning content in {\color{linecolor_green} {green}} and incorrect (or not grounded in visual evidence) reasoning content in {\color{red} {red}}. 
The proprietary model Gemini-2.5-pro does not explicitly expose its reasoning chain. Therefore, some cases only show their final answer. 
Bolded words such as ``wait", ``no", and ``but" indicate patterns of overthinking.

\input{fig/color_illusion_example}

\input{fig/motion_illusion_example}

\input{fig/gestalt_illusion_example}

\input{fig/geometric_illusion_example}

\input{fig/visual_illusion_example}

\input{fig/visual_anomaly_example}

\end{appendix}

%% file: tab/templete.tex
\begin{table*}[htbp]
\centering
\caption{Question Examples for \benchmark.}
    \label{tab:question_examples}
    \renewcommand{\arraystretch}{1.5} 
    
    \begin{tabular}{r|p{0.68\textwidth}}
    \toprule
        \textbf{Taxonomy} & \textbf{Question Example} \\
        \midrule
        Visual illusions & 
        \rowcolors{1}{lightgray}{white} 
          {\begin{tabular}[t]{@{}p{\linewidth}@{}}
            \textbullet{} \textit{How many people are on this person's hand?} \\
            \textbullet{} \textit{How many different sizes and colors of circles are in the image?} \\
            \textbullet{} \textit{Can the car get through?}\\
            \textbullet{} \textit{How many faces do you see in the image?}\\
            \textbullet{} \textit{Is the person touching the Egyptian lion in the image?}\\
            \textbullet{} \textit{Can the water flow into the cup in the boy's hand?}\\
            \textbullet{} \textit{Are there people visible in the image?}\\
            \textbullet{} \textit{Is the object symmetrical in the image?}\\
            \textbullet{} \textit{Is the person touching the Leaning Tower of Pisa in the image?}\\
            \textbullet{} \textit{Can the ball in the picture roll?}\\
            
        \end{tabular} }\\
        
        \midrule
        Geometric \& spatial illusions & 
         \rowcolors{1}{lightgray}{white} 
        {\begin{tabular}[t]{@{}p{\linewidth}@{}}
            \textbullet{} \textit{Is there a real hole in the image?} \\
            \textbullet{} \textit{Are there parallel lines in the image?} \\
            \textbullet{} \textit{Are the objects in the image aligned straight or slanted?}\\
            \textbullet{} \textit{Are the line segments near A and B equal in length?}\\
            \textbullet{} \textit{Can the figure shown in the image exist in the real three-dimensional world?}\\
            \textbullet{} \textit{Is this a three-dimensional figure?}\\
            
        \end{tabular}} \\
        
        \midrule
        Gestalt illusions & 
         \rowcolors{1}{lightgray}{white} 
         {\begin{tabular}[t]{@{}p{\linewidth}@{}}
            \textbullet{} \textit{Find the object that is different from the others, and mark its position.} \\
        \end{tabular} }\\

        \midrule
        Motion illusions & 
         \rowcolors{1}{white}{lightgray} 
        {\begin{tabular}[t]{@{}p{\linewidth}@{}}
            \textbullet{} \textit{Is the image moving or pulsing?} \\
        \end{tabular} }\\

        \midrule
        Color illusions & 
         \rowcolors{1}{lightgray}{white} 
        {\begin{tabular}[t]{@{}p{\linewidth}@{}}
            \textbullet{} \textit{Are the areas pointed to by arrows A and B the same color?} \\
            \textbullet{} \textit{What number or word is shown in the image?} \\
        \end{tabular} }\\

        \midrule
        Visual anomalies & 
         \rowcolors{1}{lightgray}{white} 
        {\begin{tabular}[t]{@{}p{\linewidth}@{}}
       
            \textbullet{} \textit{How many fingers are in the image?} \\
            \textbullet{} \textit{How many toes are in the image?} \\
        \end{tabular} }\\
        
         \bottomrule
    \end{tabular}

\end{table*}

%% file: tab/num-runs.tex
\begin{table*}[t!]
\centering
\caption{Results of each run. Because model outputs vary, results can differ slightly across runs. Therefore, we report the mean accuracy over five runs to reduce randomness.}

\label{tab:num_runs}
\setlength{\tabcolsep}{1.1pt}
\resizebox{1.0\linewidth}{!}{
\begin{tabular}{c@{\hspace{3mm}}cccccccccccc}
\toprule
\multirow{3}{*}{\begin{tabular}[l]{@{}l@{}}\textbf{Run}\end{tabular}}  &
\multicolumn{2}{c}{\rotlabel{VA}}  &
\multicolumn{2}{c}{\rotlabel{CI}} &
\multicolumn{2}{c}{\rotlabel{MI}} &
\multicolumn{2}{c}{\rotlabel{GI}} &
\multicolumn{2}{c}{\rotlabel{GSI}} &
\multicolumn{2}{c}{\rotlabel{VI}} \\
\cmidrule(lr){2-3} \cmidrule(lr){4-5} \cmidrule(lr){6-7} \cmidrule(lr){8-9} \cmidrule(lr){10-11} \cmidrule(lr){12-13}
& 
\text{Match} & \text{Judge} &
\text{Match} & \text{Judge} &
\text{Match} & \text{Judge} &
\text{Match} & \text{Judge} &
\text{Match} & \text{Judge} &
\text{Match} & \text{Judge} \\
\midrule
\rowcolor{lightgray}\multicolumn{13}{l}{\textbf{Openai o3}} \\
1 & 20.50 & 20.50 & 50.75 & 50.75 & 93.59 & 93.59 & 30.77 & 30.77 & 93.21 & 93.21 & 68.22 & 66.67 \\
2 & 27.76 & 21.76 & 51.74 & 51.74 & 91.03 & 91.03 & 35.04 & 35.04 & 92.59 & 92.59 & 66.67 & 65.89 \\
3 & 20.08 & 20.08 & 52.74 & 52.74 & 92.31 & 92.31 & 40.17 & 40.17 & 92.59 & 91.98 & 69.77 & 68.99 \\
4 & 23.85 & 23.85 & 53.73 & 53.73 & 87.18 & 87.18 & 40.17 & 40.17 & 92.59 & 92.59 & 67.44 & 67.44 \\
5 & 23.85 & 23.85 & 50.25 & 50.25 & 88.46 & 88.46 & 37.61 & 37.61 & 91.36 & 91.36 & 66.67 & 66.67 \\
\rowcolor{lightgray}\multicolumn{13}{l}{\textbf{Qwen3-VL-235B-A22B-Thinking}} \\
1 & 23.43 & 19.67 & 35.32 & 35.82 & 94.23 & 94.23 & 57.26 & 57.26 & 82.72 & 82.72 & 62.02 & 66.67 \\
2 & 23.01 & 19.25 & 36.82 & 34.83 & 94.87 & 94.87 & 56.41 & 55.56 & 82.10 & 81.48 & 62.79 & 65.89 \\
3 & 23.85 & 20.50 & 39.30 & 37.81 & 95.51 & 95.51 & 59.83 & 54.70 & 87.04 & 86.42 & 63.57 & 65.89 \\
4 & 23.43 & 20.08 & 38.81 & 37,31 & 94.87 & 94.87 & 58.12 & 57.26 & 80.25 & 80.25 & 62.02 & 64.34 \\
5 & 23.85 & 20.92 & 37.81 & 36.32 & 98.08 & 98.08 & 59.83 & 57.26 & 85.80 & 83.95 & 61.24 & 65.12 \\
\bottomrule
\end{tabular}
}
\end{table*}

%% file: tab/model_version.tex
\begin{table*}[htbp]
\centering
\caption{Model markers or links used in our evaluation.}
\label{tab:model_version}
\setlength{\tabcolsep}{1.2pt}
\resizebox{0.95\linewidth}{!}{
\begin{tabular}{>{\small}lc}
\toprule
\textbf{Model} & \textbf{Marker/Link}\\
\midrule
Gemini-3-pro & \em api\_google\_gemini-3-pro-preview\\

\rowcolor{lightgray}GPT-4-Turbo & \em api\_openai\_gpt-4-turbo-2024-04-09 \\

GPT-4.1-mini & \em api\_openai\_gpt-4.1-mini \\

\rowcolor{lightgray} Gemini-2.5-pro & {\em api\_google\_gemini-2.5-pro}\\

GPT-5-chat-latest & \em api\_openai\_gpt-5-chat-latest-response\\

\rowcolor{lightgray} GPT-4o-2024-11-20 & \em api\_azure\_openai\_gpt-4o-2024-11-20\\

ChatGPT-4o-latest & \em api\_openai\_chatgpt-4o-latest\\

\rowcolor{lightgray} OpenAI o4-mini & \em api\_azure\_openai\_o4-mini\\

GPT-5-mini & \em api\_azure\_openai\_gpt-5-mini-response \\

\rowcolor{lightgray} Qwen-vl-max-latest & api\_ali\_qwen-vl-max-latest\\


OpenAI o3 & \em api\_openai\_o3\\

\rowcolor{lightgray} GLM-4.5V & \em api\_zhipu\_glm-4.5v \\

Claude-3.5-sonnet-20241022 & \em api\_anthropic\_claude-3-5-sonnet-20241022 \\

\rowcolor{lightgray} Claude-sonnet-4-20250514 & \em api\_anthropic\_claude-sonnet-4-20250514 \\

Claude-opus-4.1-20250805 & \em api\_anthropic\_claude-opus-4-1-20250805\\

\rowcolor{lightgray} Qwen2.5-VL & \em \url{https://github.com/QwenLM/Qwen2.5-VL}\\

InternVL3.5 & \em \url{https://github.com/OpenGVLab/InternVL}\\

\rowcolor{lightgray} Qwen3-VL & \em \url{https://github.com/QwenLM/Qwen3-VL}\\
\bottomrule
\end{tabular}
}
\end{table*}

%% file: fig/system_prompt.tex
\begin{figure*}
    \centering
    \includegraphics[width=1.0\linewidth]{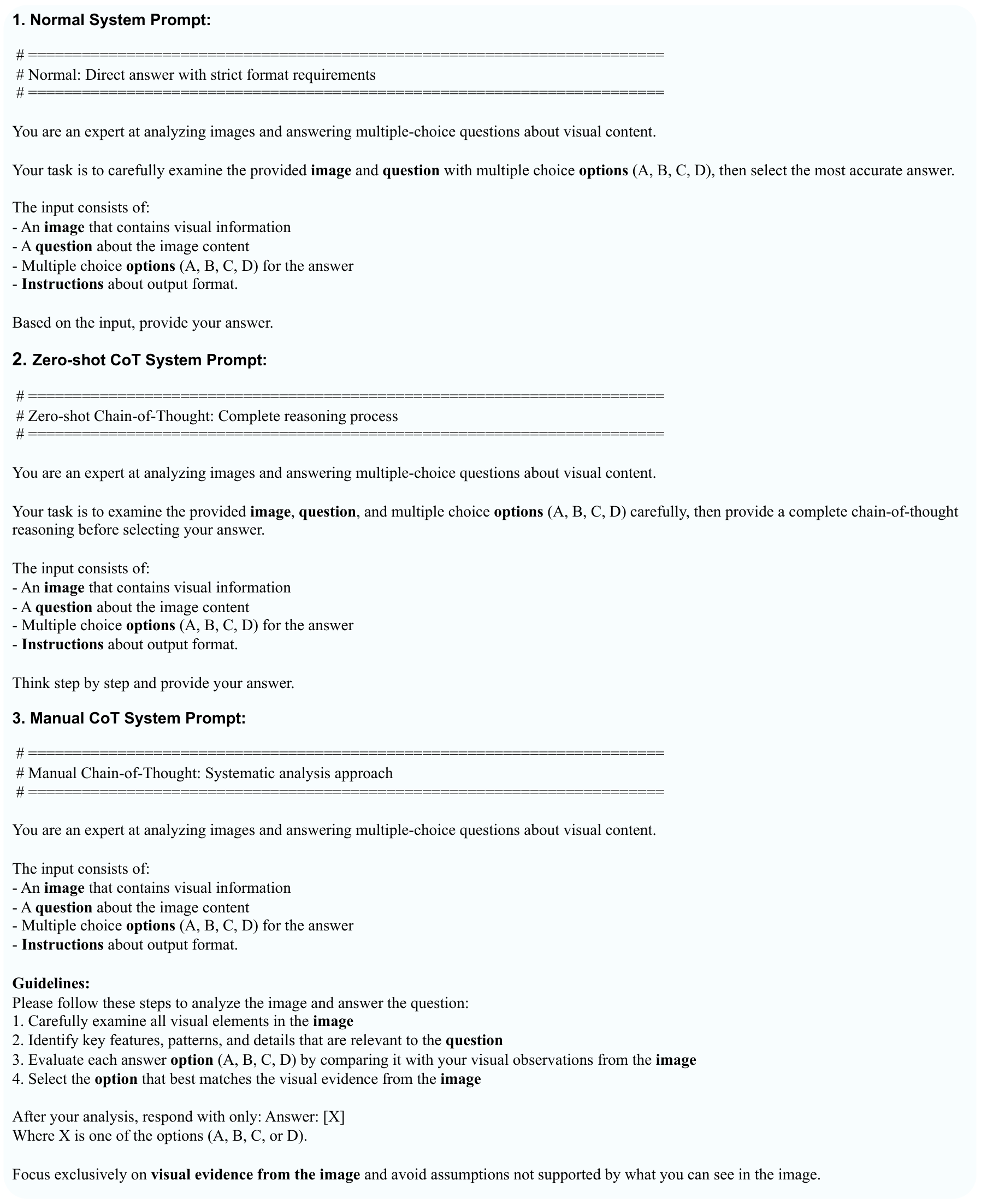}
    \caption{Illustration of three system prompts adopted in \benchmark evaluation.}
    \label{fig:system_prompt}
\end{figure*}

%% file: fig/color_illusion_example.tex
\begin{figure*}
    \centering
    \includegraphics[width=1.0\linewidth]{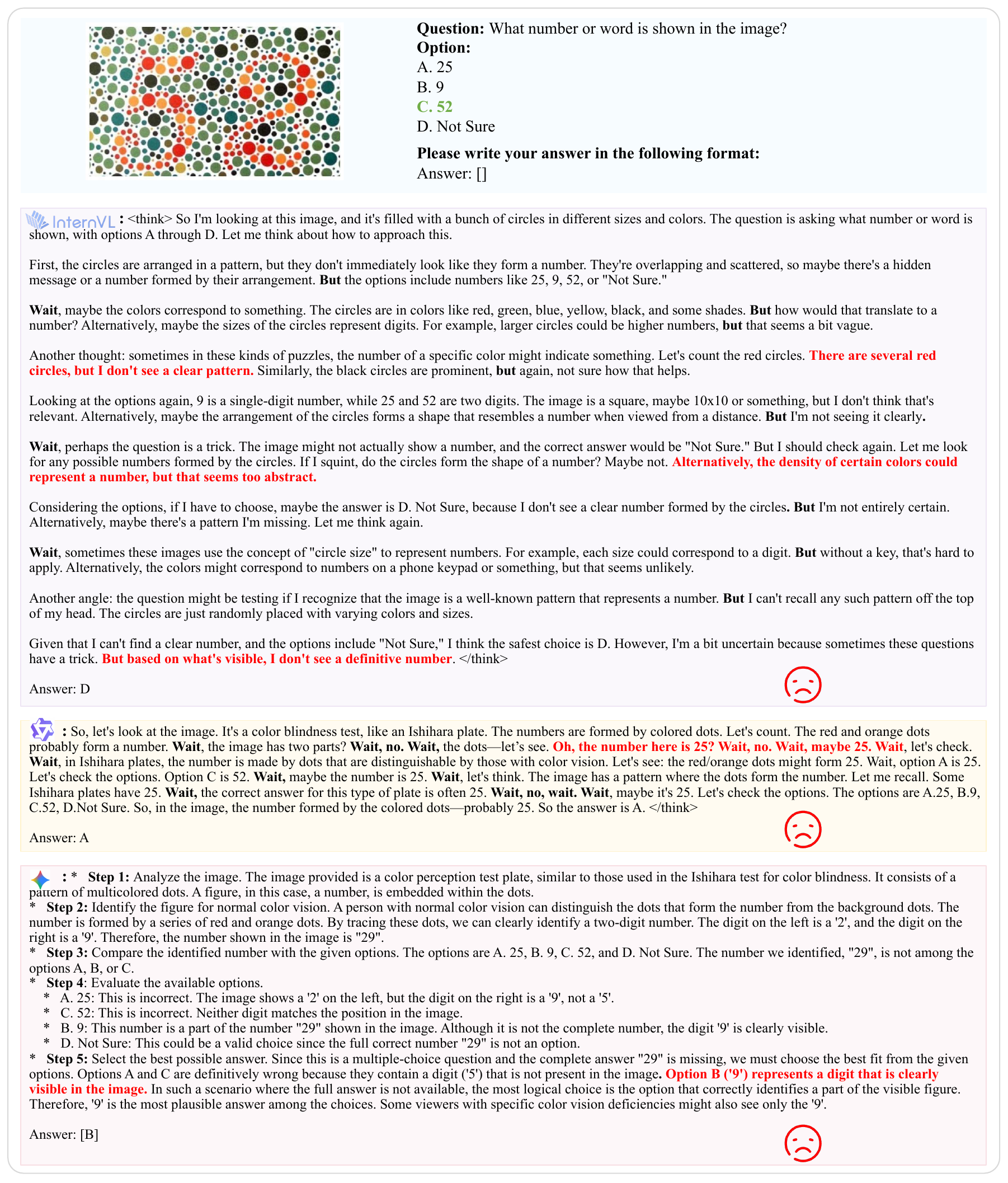}
    \caption{A demonstration of color illusions. InternVL-3.5-8B does not see a clear number and exposes a severe phenomenon of overthinking. 
    Qwen3-VL-30B-A3B-Thinking identifies a reversed order and exhibits very low confidence in its answer, which leads to repetitive confirmation.
    Gemni-2.5-pro makes a recognition error, yet it preserves a readable and well-structured reasoning process.}
    
    \label{fig:color_illusion_example}
\end{figure*}

%% file: fig/motion_illusion_example.tex
\begin{figure*}[htbp]
    \centering
    \includegraphics[width=1.0\linewidth]{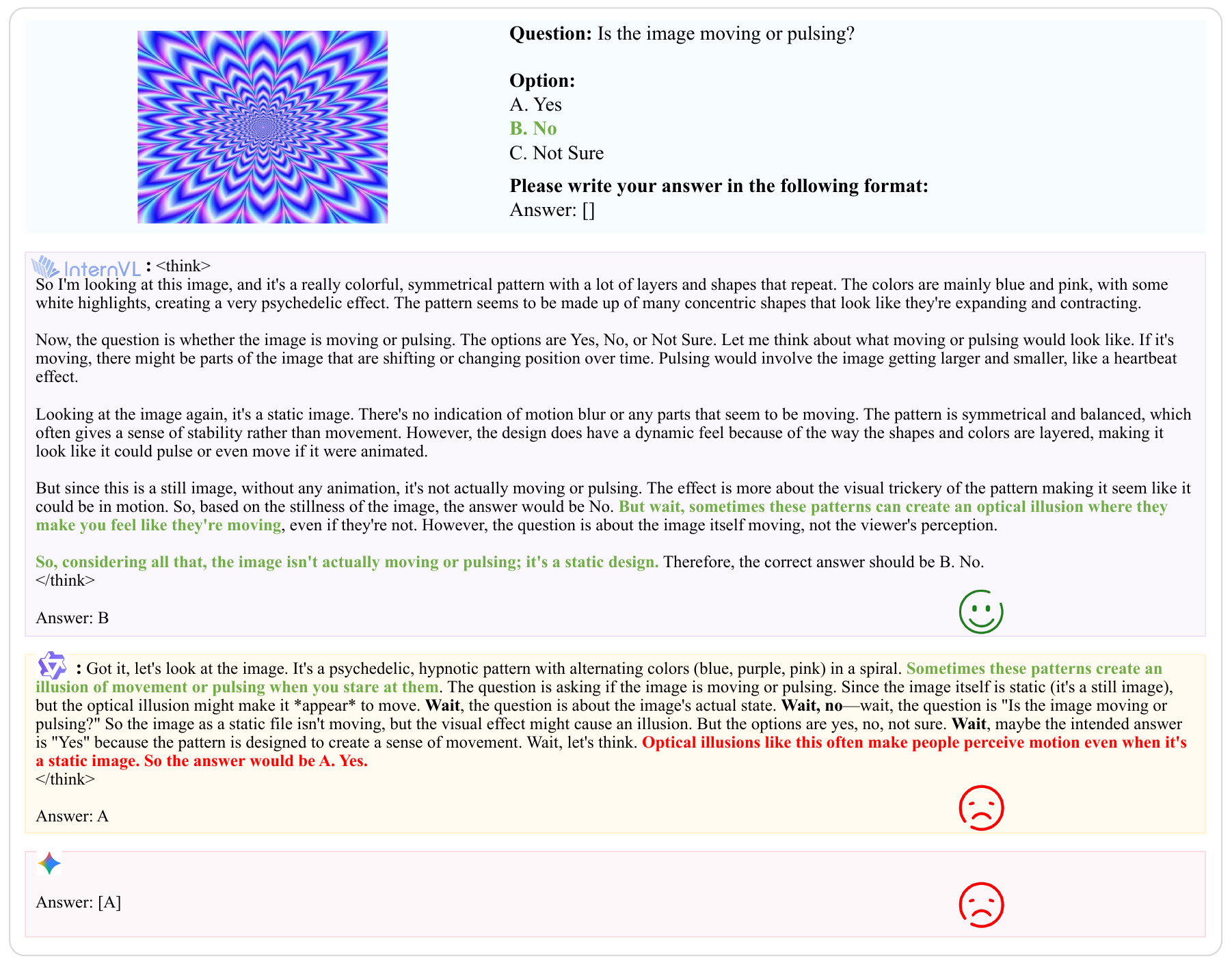}
    \caption{A demonstration of motion illusions.
    InternVL-3.5-8B identifies this as an optical illusion and infers from the visual content that it is a static image. 
    Qwen3-VL-30B-A3B-Thinking recognizes the illusion, but its overthinking leads to an incorrect answer. 
    Gemni-2.5-pro directly answers ``Yes''.}

    \label{fig:motion_illusion_example}
\end{figure*}

%% file: fig/gestalt_illusion_example.tex
\begin{figure*}[htbp]
    \centering
    \includegraphics[width=0.99\linewidth]{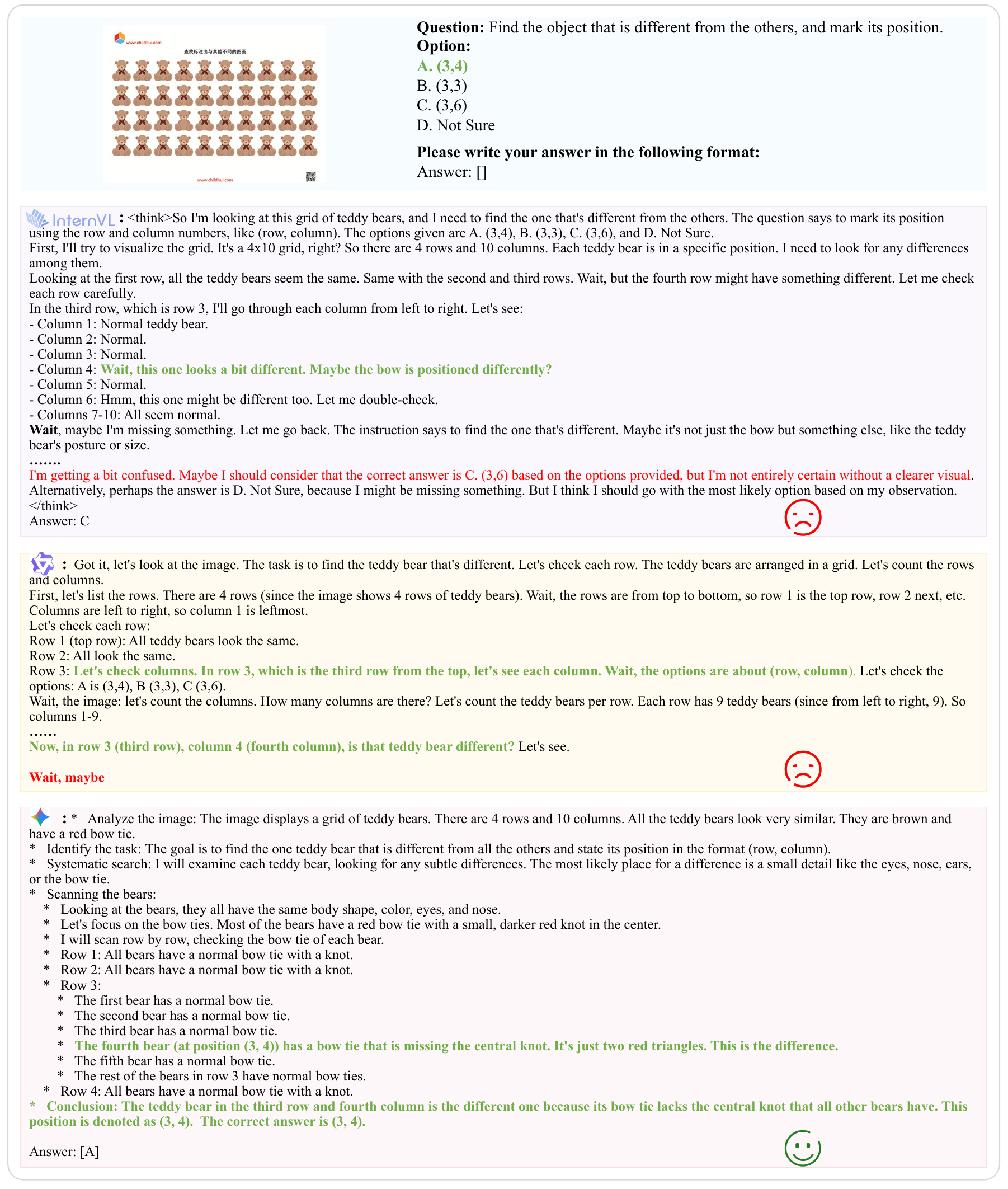}
    \vspace{-3mm}
    \caption{A demonstration of gestalt illusions.
     The ellipsis (``......'') denotes the omission of intermediate reasoning steps. 
     In gestalt illusions, models must accurately perceive fine-grained details and infer their precise locations. InternVL-3.5-8B identifies the discrepancy in the fourth column but remains unconfident, eventually selecting the incorrect answer after analyzing the options. Although Qwen3-VL-30B-A3B-Thinking recognizes the difference, its iterative reasoning exceeds the maximum output length, failing to produce a direct answer. In contrast, Gemini-2.5-pro demonstrates effective pattern recognition, resulting in the correct outcome.}
    \label{fig:gestalt_illusion_example}
\end{figure*}

%% file: fig/geometric_illusion_example.tex
\begin{figure*}
    \centering
    \includegraphics[width=1.0\linewidth]{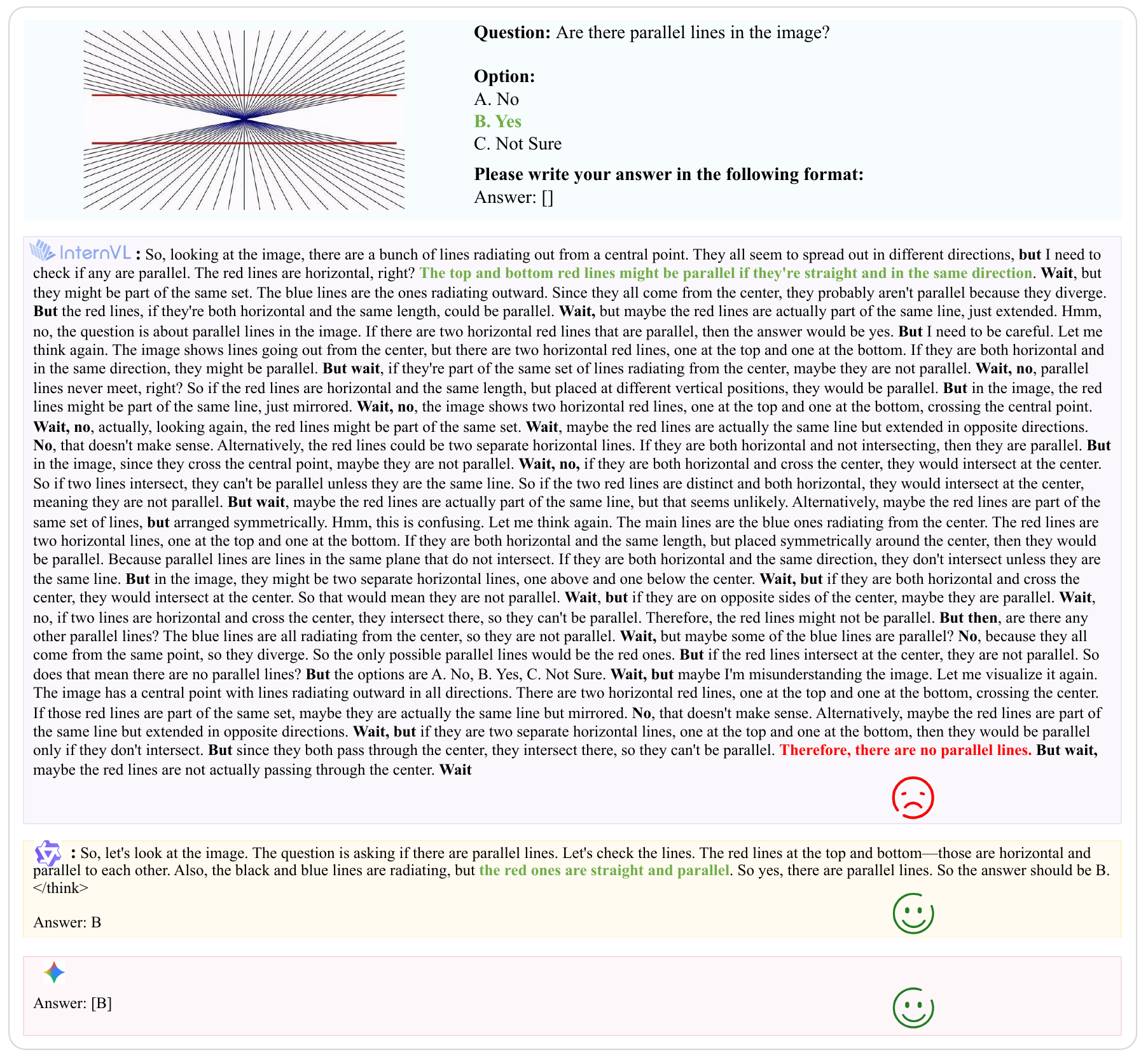}
    \caption{A demonstration of geometric and spatial illusions.
    In this case, Qwen3-VL-30B-A3B-Thinking and Gemini-2.5-pro concisely arrive at the correct choice. 
    InternVL-3.5-8B initially suggests the red lines might be parallel. However, it subsequently gets stuck repeating this viewpoint, which leads to an incorrect conclusion. Ultimately, it fails to output a final choice because its judgment exceeds the maximum output length.}
    \label{fig:geometric_illusion_example}
\end{figure*}

%% file: fig/visual_illusion_example.tex
\begin{figure*}
    \centering
    \includegraphics[width=1.0\linewidth]{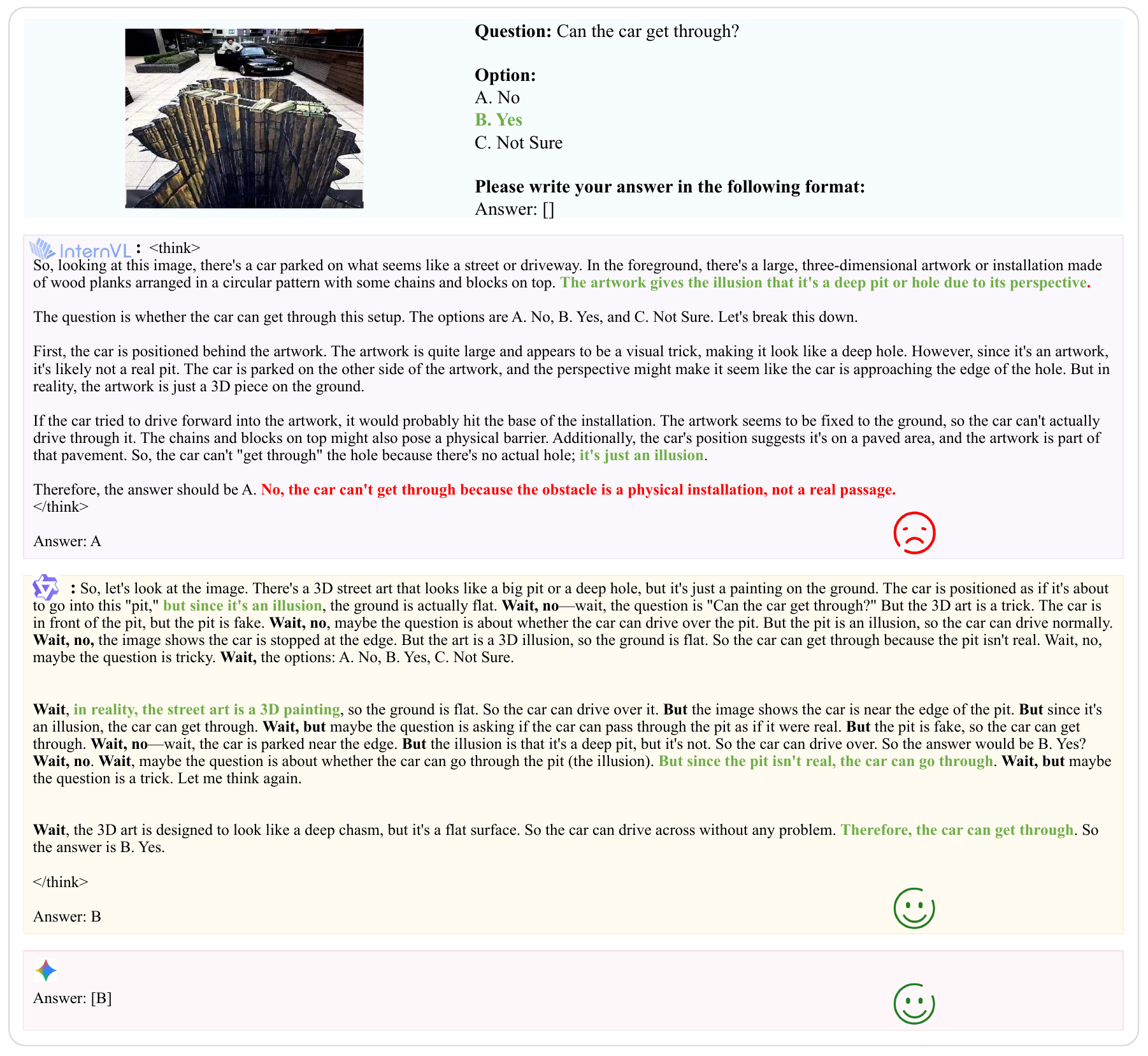}
    \caption{A demonstration of visual illusions.
    Both InternVL-3.5-8B and Qwen3-VL-30B-A3B-Thinking argue that the image content is an illusion. 
    However, InternVL-3.5-8B follows an incorrect reasoning path. 
    In contrast, Qwen3-VL-30B-A3B-Thinking, despite excessive reasoning, correctly concludes that the car can get through.
    }
    \label{fig:visual_illusion_example}
\end{figure*}

%% file: fig/visual_anomaly_example.tex
\begin{figure*}
    \centering
    \includegraphics[width=1.0\linewidth]{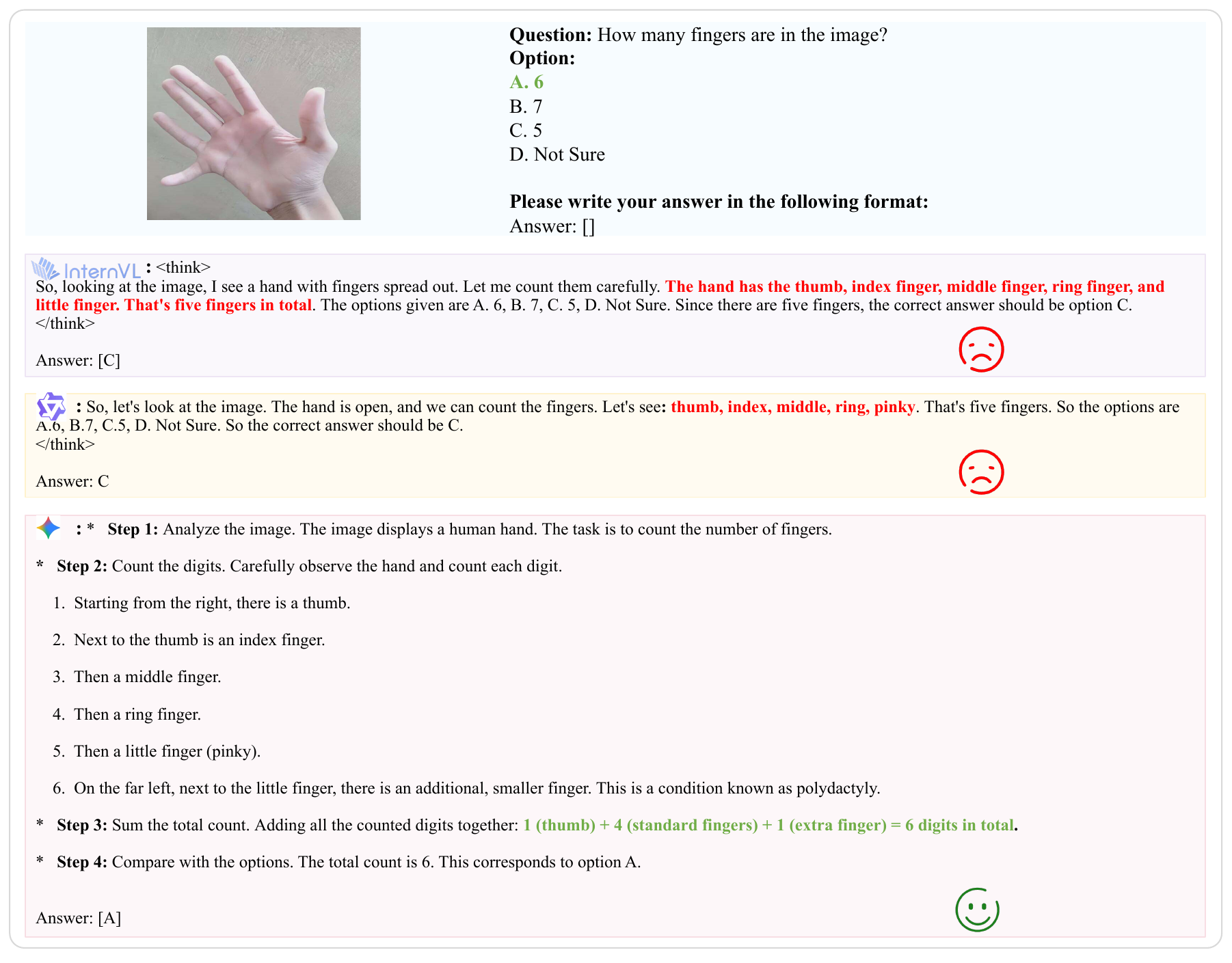}
    \caption{A demonstration of visual anomalies.
    Visual anomalies that violate commonsense priors, such as biological irregularities (\eg, polydactyly), remain challenging for current MLLMs.
    As illustrated, InternVL-3.5-8B and Qwen3-VL-30B-A3B-Thinking rely on a ``five-finger'' prior and miscount, indicating insufficient grounding in the image.
    By contrast, Gemini-2.5-pro first identifies each finger, then summarizes and counts them, yielding the correct prediction.
    }
    \label{fig:visual_anomaly_example}
\end{figure*}